\documentclass[11pt,table,dvipsnames]{article}

\usepackage[final]{neurips_2025}
\usepackage{times}
\usepackage{algorithmic}
\usepackage{algorithm}
\usepackage{amsmath}
\usepackage{latexsym}
\usepackage[T1]{fontenc}

\usepackage[utf8]{inputenc}

\usepackage{inconsolata}

\usepackage{graphicx}
\title{The Lighthouse of Language: Enhancing LLM Agents via Critique-Guided Improvement}

\author{
 Ruihan Yang$^{\heartsuit}$\thanks{Work done during an internship at Tencent Hunyuan.}, 
 Fanghua Ye$^{\spadesuit}$\thanks{Corresponding authors.}, 
 Jian Li$^{\spadesuit}$\footnotemark[2], 
 Siyu Yuan$^{\heartsuit}$,
 Yikai Zhang$^{\heartsuit}$,  \\
 \textbf{Zhaopeng Tu}$^{\spadesuit}$,
 \textbf{Xiaolong Li}$^{\spadesuit}$,
 \textbf{Deqing Yang}$^{\heartsuit}$\footnotemark[2]\\
 $^{\heartsuit}$Fudan University \quad 
 $^\spadesuit$Tencent Hunyuan
 \\
    \small $^\heartsuit$\texttt{\{rhyang17\}@fudan.edu.cn} \quad
    \small $^\spadesuit$\texttt{\{fanghua.ye.21,lijianjack\}@gmail.com}\\
    \quad \\
    Project Page: \url{https://github.com/rhyang2021/CGI}
}

\usepackage{microtype}
\usepackage{graphicx}
\usepackage{arydshln}
\usepackage{inconsolata}
\usepackage{todonotes}
\usepackage{tabularx}
\usepackage{booktabs} 
\usepackage{hyperref}
\usepackage{url}
\usepackage{amsfonts}
\usepackage{nicefrac}
\usepackage{multirow}
\usepackage{amssymb}
\usepackage{subcaption}
\usepackage{fge}

\usepackage[framemethod=TikZ]{mdframed} 
\usepackage{listings}
\usepackage{longtable}
\usepackage{tcolorbox}
\usepackage{xspace}
\usepackage{wrapfig}
\usepackage{pgfplots}
\usepackage{color, colortbl}
\usepackage{array}
\usepackage{paralist}
\usepackage{geometry}
\usepackage{enumitem}
\usepackage{natbib}
\setcitestyle{numbers,square}
\usepackage[nameinlink,capitalise]{cleveref}
\hypersetup{
    colorlinks,
    linkcolor={blue!80!black},
    citecolor={blue!80!black},
}

\definecolor{Gray}{gray}{0.92}
\definecolor{racing-green}{rgb}{0.0, 0.8, 0.6}
\definecolor{awesome-red}{rgb}{1.0, 0.13, 0.32}
\definecolor{LightCyan}{rgb}{0.88,1,1}
\definecolor{darkgreen}{RGB}{0,150,0}
\definecolor{Ground}{RGB}{255,184,55}
\definecolor{Dirt}{RGB}{191,169,115}
\definecolor{Pink}{RGB}{226,184,176}
\definecolor{Violet}{RGB}{163,148,170}
\definecolor{darkred}{RGB}{150,0,0} %

\definecolor{lightblue}{RGB}{208,227,251}
\definecolor{level4}{RGB}{110,136,203}
\definecolor{level3}{RGB}{173,190,226}
\definecolor{level2}{RGB}{205,208,243}
\definecolor{level1}{RGB}{236,236,252}

\newcolumntype{g}{>{\columncolor{Ground!7}}c}
\newcolumntype{d}{>{\columncolor{cyan!6}}c}
\newcolumntype{f}{>{\columncolor{lime!6}}c}
\newcolumntype{v}{>{\columncolor{purple!6}}c}

\newcommand{\ie}{\textit{i}.\textit{e}.,\ }
\newcommand{\eg}{\textit{e}.\textit{g}.,\ }
\newcommand{\method}{\textsc{CGI}\xspace}

\begin{document}
\maketitle
\begin{abstract}
Large language models (LLMs) have recently transformed from text-based assistants to autonomous agents capable of planning, reasoning, and iteratively improving their actions. 
While numerical reward signals and verifiers can effectively rank candidate actions, they often provide limited contextual guidance. 
In contrast, natural language feedback better aligns with the generative capabilities of LLMs, providing richer and more actionable suggestions. However, parsing and implementing this feedback effectively can be challenging for LLM-based agents. 
In this work, we introduce \textbf{C}ritique-\textbf{G}uided \textbf{I}mprovement (\method), a novel two-player framework, comprising an actor model that explores an environment and a critic model that generates detailed nature language feedback. 
By training the critic to produce fine-grained assessments and actionable revisions, and the actor to utilize these critiques, our approach promotes more robust exploration of alternative strategies.
Experiments in three interactive environments show that \method outperforms existing baselines by a substantial margin. 
Notably, even a small critic model surpasses GPT-4 in feedback quality. 
The resulting actor achieves state-of-the-art performance, demonstrating the power of explicit guidance to enhance decision-making in LLM-based agents.
\end{abstract}

\section{Introduction}

Recent advances in large language models (LLMs)~\citep{openai2024gpt4technicalreport, Gemini2023, llama3modelcard} have broadened their applicability from conventional chat and text-based interactions to more dynamic, agentic roles. 
In these roles, LLM-based agents~\citep{durante2024agentaisurveyinghorizons,chen2024personapersonalizationsurveyroleplaying,li2024selu} harness the reasoning and planning capabilities of LLMs to automate complex tasks across diverse domains, including code generation~\citep{yuan2024easytoolenhancingllmbasedagents,qin2023toolllmfacilitatinglargelanguage}, software engineering~\citep{hu2025selfevolving,jimenez2024swebench}, and web applications~\citep{yao2023webshop,putta2024agentqadvancedreasoning}. 
A key requirement in these agentic tasks is the ability to iteratively acquire, store, and use new information to improve performance. 
Thus, a central challenge emerges: \textit{How can high-quality feedback be obtained and utilized efficiently?}  

A common approach for providing feedback is to rely on numerical signals, such as verifiers~\citep{wang-etal-2024-math,qian2025discriminatorguided} or reward models~\citep{xia2025agentrmenhancingagentgeneralization,chen2025autonomous}. 
In this paradigm, LLMs are typically trained as discriminators to evaluate and select the best action from a set of candidates. 
One widely used method is Best-of-$N$ (BoN)~\citep{stiennon2022learningsummarizehumanfeedback, ouyang2022training}, where the agent generates $N$ possible actions and a trained reward model picks the highest-scoring one. 
However, numerical feedback has limited informativeness: it reduces guidance to scalar scores that merely filter candidate actions, overlooking context-specific suggestions, avenues for exploration, or strategies for refining suboptimal behavior.

A more direct and flexible approach involves using natural language feedback. 
Rather than providing only a ``best'' action, LLMs can generate detailed critiques and explicit, context-aware recommendations for improvement. 
However, transitioning to natural language feedback presents two main challenges:
\begin{inparaenum}[\it 1)]  
\item \emph{Weak Feedback}: Many techniques rely on self-refinement or self-correction~\citep{madaan2024self,shinn2023reflexion,yang2024selfgoal}, which heavily depend on the LLM’s inherent capacity. This reliance can lead to degraded performance, especially when models hallucinate or encounter complex problems~\citep{shinn2023reflexion,kumar2025training};
\item \emph{Poor Utilization}: Unlike numerical feedback, which directly selects an action to execute, verbal critiques require the agent to accurately interpret and integrate the guidance into its decision-making process. LLM agents often struggle with this, sometimes demonstrating limited flexibility in adjusting or discarding previously generated plans~\citep{cuadron2025dangeroverthinkingexaminingreasoningaction,metz2025reward}.
\end{inparaenum}

In response, we propose \textbf{C}ritique-\textbf{G}uided \textbf{I}mprovement (\method), a two-player framework in which an actor model interacts with the environment while a critic model provides supervisory feedback. \method features two main stages: \textbf{Critique Generation} and \textbf{Action Refinement}. 
In the Critique Generation stage, we train a critic model to produce precise evaluations and actionable revision instructions. 
In the Action Refinement stage, the actor model learns to effectively apply these critiques through iterative supervised fine-tuning, thereby improving both its reasoning and its ability to integrate external feedback.

We conduct extensive experiments in three interactive environments~\citep{yao2023webshop,wang2022scienceworlda,prasad2024}, demonstrating that \method substantially enhances performance over baseline approaches. Notably, our critic model, when trained on a small backbone (e.g., Llama-3-8B), outperforms even GPT-4 as a critic by a large margin (+29.16\%). Additionally, the action-refinement process further boosts results, surpassing state-of-the-art metrics by 26.74\%.

In summary, our contributions are as follows:  
\begin{inparaenum}[\it 1)]  
    \item We tackle the challenge of obtaining and utilizing high-quality nature language feedback in agentic tasks.
    \item We present Critique-Guided Improvement (\method), a novel two-player framework that provides more informative feedback to LLM-based agents by combining a specialized critic model with an actor model.
    \item We develop a training strategy that iteratively refines the actor's actions based on natural language critiques, enabling both improved reasoning and enhanced adaptability. 
    \item We conduct extensive experiments across three diverse interactive environments. By leveraging \method, the agent can continuously improve its performance in long-horizon tasks and enhance task efficiency by achieving higher scores in fewer steps.
\end{inparaenum}

\section{Related Work}
\paragraph{Learning from Feedback}
Current feedback can be categorized into numerical and natural language feedback. Numerical feedback~\citep{wang-etal-2024-reward-modeling, cobbe2021trainingverifierssolvemath,yu-etal-2024-ovm,luo2024improvemathematicalreasoninglanguage} is typically provided by training a reward model (RM) or verifiers. 
In this approach, the model learns to predict either the correctness of a solution~\citep{wang-etal-2024-math,yu-etal-2024-ovm} or the preference between multiple solutions~\citep{stiennon2022learningsummarizehumanfeedback,nakano2022webgptbrowserassistedquestionansweringhuman}. Specifically, the RM generates a continuous numerical score, which is then incorporated into a classification objective. 
In contrast, natural language feedback offers denser rewards by evaluating the model's actions using natural language. This feedback can be generated in two main ways: by prompting an off-the-shelf LLM to act as a verifier (\ie LLM-as-judge)
~\citep{bai2022constitutionalaiharmlessnessai,zheng2023judgingllmasajudgemtbenchchatbot}, or through self-refinement and self-critique~\citep{shinn2023reflexion,madaan2023selfrefine}, where the model evaluates and improves its own outputs. Our trained critic model provides natural language feedback, offering denser and more accurate rewards.

\paragraph{Agent Learning in Interactive Environments.}
Previous approaches to agent learning in interactive environments can be classified into three main categories:
\begin{inparaenum}[\it 1)]
\item Prompt-based methods~\citep{yang2024selfgoal, yao2023react, shang2025agentsquareautomaticllmagent, chen2024automanualconstructinginstructionmanuals} utilize human-written prompts to guide LLMs in summarizing experiences. These summaries, which may include causal abstractions from both successful and failed attempts~\citep{shinn2023reflexion, majumder2023clincontinuallylearninglanguage} or transferable skills~\citep{nottingham2024sso}, are integrated into the model’s memory to enhance its knowledge and performance.
\item Training-based methods~\citep{xi2024agentgymevolvinglargelanguage, putta2024agentqadvancedreasoning, yuan2025agentrtraininglanguagemodel} rely on techniques such as Supervised Fine-Tuning (SFT)~\citep{chen-etal-2024-agent, song-etal-2024-agentbank} or Direct Preference Optimization (DPO)~\citep{shi-etal-2024-direct, song2024trialerrorexplorationbasedtrajectory}.
to train LLMs. The training data may come from expert models or be generated through exploration strategies like Monte Carlo Tree Search (MCTS).
\item Inference-time sampling methods~\citep{qian2025discriminatorguided, xia2025agentrmenhancingagentgeneralization, chen2025autonomous} employ techniques such as Best-of-N (BoN)~\citep{stiennon2022learningsummarizehumanfeedback, ouyang2022training} and Tree-of-Thought (ToT)~\citep{yao2023tree} 
to identify optimal actions during inference. These methods leverage the prior knowledge in LLMs, enabling more efficient search processes. Our \method method introduces a novel variation of inference-time sampling by using a trained critic that evaluates candidate actions and suggests revisions to enhance inference performance.
\end{inparaenum}

\section{Preliminary}
\paragraph{Partially Observable Markov Decision Process}
We define the collection of environments as $\mathcal{E}$.
For a specific environment $e \in \mathcal{E}$, 
the embodied tasks for LLM agents are typically modeled as a Partially Observable Markov Decision Process (POMDP): $(\mathcal{X}, \mathcal{S}, \mathcal{A}, \mathcal{O}, \mathcal{T})_e$.  
Here, $\mathcal{X}$ denotes the set of instructions,  
$\mathcal{S}$ represents the set of environment states,  
$\mathcal{A}$ is the set of available actions at each state, and  
$\mathcal{O}$ represents the observations available to the agent.  
The transition function $\mathcal{T}: \mathcal{S} \times \mathcal{A} \rightarrow \mathcal{S}$ is defined by the environment, while the reward function $\mathcal{R}: \mathcal{S} \times \mathcal{A} \rightarrow [0, 1]$
specifies the reward received by the agent from the environment.  
For an actor model $\pi_{\theta}$ parameterized by $\theta$, the actor selects the next action $a_t \sim \pi_{\theta}(a | \tau_t, e)$ at each time step $t$, based on the interaction history $\tau_t$, which is defined as  
$$
\tau_t = \left\{x, a_0, o_0, \ldots, a_{t-1}, o_{t-1}\right\}, \quad \tau_0 = \left\{x\right\}.
$$ 
The trajectory is then represented as 
$$
\tau=\left(x, a_0, o_0, \ldots, a_{T}, o_T\right) \sim \pi_\theta(\tau \mid x, e).
$$


\paragraph{Iterative Supervised Fine-Tuning in Agentic Task}

Iterative Supervised Fine-Tuning (SFT) is a process of exploration and learning~\citep{aksitov2023restmeetsreactselfimprovement,huang-etal-2023-large, qu2024recursiveintrospectionteachinglanguage}. It iteratively utilizes the correct responses from the actor model to enhance the model's problem-solving abilities. 
The process involves \(K\) iterations, each consisting of two steps: exploration and learning.
For the collection of environments $\mathcal{E}$, in the exploration step of iteration \(k\), the model \(\pi_\theta^{k-1}\) from the previous iteration is applied to each environment $e$, resulting in \(\vert \mathcal{E}\vert\) trajectories \(\{\tau^{(j)}\}_{j=1}^{\vert \mathcal{E}\vert}\). These trajectories are then filtered using the reward function \(\mathcal{R}(\tau)\), retaining only the correct ones. The filtered trajectories form a new dataset \(\mathcal{D}_{\text{correct}} = \{ \tau^{(j)}\}_{j=1}^{N^{\prime}}\), where \(N'\) is the number of trajectories retained. In the learning step of iteration \(k\), this new dataset is used to fine-tune the actor model \(\pi_\theta\) to obtain \(\pi^k_\theta\).

\section{Methodology}  
\begin{figure*}[t]
    \centering
    \includegraphics[width=0.99\linewidth]{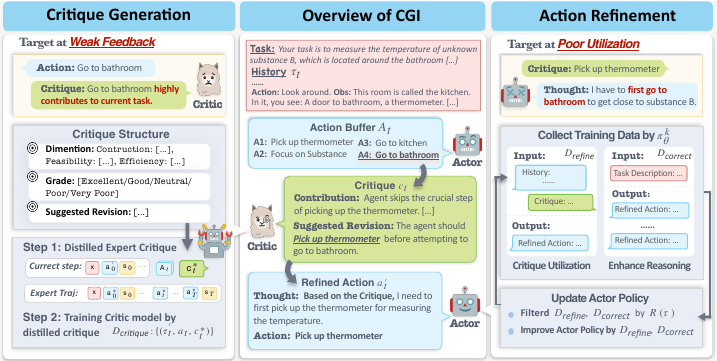}
    \caption{An overview of \method, illustrated with a ScienceWorld example. The actor first generates candidate actions, and the critic then assesses each action and generates suggested revisions. The actor refines its actions based on the critique. The \textbf{Critique Generation} stage enhances the critic model's ability to provide effective feedback, while the \textbf{Action Refinement} stage helps the actor model better utilize the critique.}
    \label{fig:workflow}
\end{figure*}
In this section, we first provide an overview of the proposed Critique-Guided Improvement (\method) framework (\S~\ref{sec:critic_guided_inference}), explaining how the actor and critic collaborate to enhance performance.  
We then detail the two key stages of \method, critique generation and action refinement, which respectively address the challenges of critic's \textit{weak feedback} and the actor's \textit{poor utilization}.  
In the \textbf{Critique Generation} stage (\S~\ref{sec:critic_generation}), the critic model learns to evaluate the actor’s candidate actions and provide actionable feedback.  
In the \textbf{Action Refinement} stage (\S~\ref{sec:critic_utilization}), the actor refines its actions by integrating critiques from the critic model into its interaction with the environment.  
Algorithm~\ref{algo:1} summarizes the \method framework (see Appendix~\ref{app:notification} for definitions of all notifications).

\subsection{Overview of the \method Framework} \label{sec:critic_guided_inference}  
The \method framework follows a two-player setting, where the actor model \(\pi_{\theta}\) generates multiple candidate actions, and the critic model \(\pi_{\phi}\) provides feedback to refine them.  
For a specific environment $e \in \mathcal{E}$, at each time step \(t\), given the refined trajectory history  
\[
\tau'_t = \{x, a_0^{\prime}, o_0, \dots, a_{t-1}^{\prime}, o_{t-1}\},
\]  
the actor generates \(M\) candidate actions, stored in the action buffer \(A_t\!=\!\{a_{t,i}\}_{i=1}^{M}\).
The critic then evaluates each candidate and generates corresponding critiques \(c_t\!=\!\{\pi_{\phi}(c \mid \tau'_{t}, a_{t,i}, e)\}_{i=1}^{M}\). Using these critiques, the actor refines its decision and generates the final action \(a^{\prime}_t\!=\!\pi_{\theta}(a \mid \tau'_{t}, c_t, e)\),  
which is then executed in the environment.  
This iterative process produces the refined trajectory:  
\[
\tau^{\prime} = \{x, a^{\prime}_0, o_0, \dots, a^{\prime}_T, o_T\}.
\]  
In this collaborative framework, higher-quality critiques \(c_t\) and better utilization of feedback lead to improved refined actions \(a_t^{\prime}\), ultimately enhancing the final performance \(\mathcal{R}(\tau^{\prime})\).  

\begin{algorithm}[t]
\caption{Critique-Guided Improvement (\method)}
\begin{algorithmic}[1]
\STATE \textbf{Input:} Initialized policy actor model $\pi_{\theta}$, 
critic model $\pi_{\phi}$,
environment set $\mathcal{E}$,
full instruction set $\mathcal{X}$, reward function $\mathcal{R}$, the expert critiques $\mathcal{D_{\rm{critique}}}$, the expert path $\mathcal{D_{\rm{expert}}}$, the general dataset $\mathcal{D_{\rm{general}}}$, max time step $T$, and the iteration rounds $K$.
\vspace{0.1cm}
\STATE \textbf{Procedure Fine-tune the  Critic Model:}
\STATE  \hspace{1em} Minimize the following objective to obtain the critic model $\mathcal{\pi}_{\phi}$:\\ 
\hspace{2em} $\mathcal{L}_{\text{critic}}(\phi) =  
    \mathbb{E}_{(c_{t}, \tau_{t}^{\prime}, a_{t}, e) \sim \mathcal{D}_{\text{critique}}} \Big[ \log \pi_\phi\left(c_{t} \mid \tau_{t}^{\prime}, a_{t}, e\right) \Big]$;
\vspace{0.1cm}
\STATE \textbf{Procedure Iterative Action Refinement:}
\STATE \hspace{1em} $\pi^{0}_{\theta} \leftarrow \pi_{\theta_{base}}$;
\FOR{iteration $k = 1$ to $K$}
    \STATE $D_{\rm{train}} \leftarrow D_{\rm{expert}}$; \textcolor{gray!60}{// Initialize $\mathcal{D}_{\rm{train}}$ with $D_{\rm{expert}}$.}
    \vspace{0.1cm}
    \STATE \textbf{Perform} 
    \textcolor{darkgreen}{Exploration Step}\\
    \hspace{1em} Actor $\pi^{k-1}_{\theta}$ interacts with $e \in \mathcal{E}$ under the guidance of critic $\pi_{\phi}$;
    \STATE \hspace{1em} \textcolor{gray!60}{// Collect critique-action pairs with $\mathcal{R}(\tau^{\prime}) = 1$.}
    \STATE \hspace{1em} Collect $D_{\rm{refine}} = \bigcup_{e \in \mathcal{E}} D_{\rm{refine}}^e$, where $D_{\rm{refine}}^e = \{(\tau_{t}^{\prime}, c_t, a'_t)\}_{t=1}^{T}$;
    \vspace{0.1cm}
    \STATE \hspace{1em} \textcolor{gray!60}{// Update training set with correct trajectory.}
    \STATE \hspace{1em} Collect $D_{\rm{correct}}=\bigcup_{e \in \mathcal{E}}D_{\rm{correct}}^e$,  $D_{\rm{train}} \leftarrow D_{\rm{train}} \cup \mathcal{D}_{\rm{correct}}$;
    \vspace{0.1cm}
    \STATE \textbf{Perform}
    \textcolor{darkred}{Learning Step}
    \vspace{0.1cm}
    \STATE \hspace{1em} Minimize the following objective to obtain actor model $\pi_{\theta}^{k}$:\\
    $\mathcal{L}_{\text{actor}}(\theta) = \beta  \ \left\{ \mathbb{E}_{(\tau, x, e) \sim \mathcal{D}_{\text{train}}} \Big[\log \pi_\theta(\tau \mid x, e)\Big] + \mathbb{E}_{(a'_t,\tau'_{t}, c_t, e) \sim \mathcal{D}_{\text{refine}}} \Big[\log \pi_\theta(a'_t \mid \tau_{t}^{\prime}, c_t, e)\Big] \right\}$ \\
    \hspace{2em} $+ (1-\beta) \ \mathbb{E}_{(x, y) \sim \mathcal{D}_{\text{general}}} \Big[\log \pi_\theta(y \mid x)\Big]
    $;
\ENDFOR
\end{algorithmic}
\label{algo:1}
\end{algorithm}
\subsection{Critique Generation}  
\label{sec:critic_generation}
The \textbf{Critique Generation} stage trains the critic model to assess the alignment between LLM-generated actions and optimal ones at each step, and convert this assessment into structured natural language critiques.  
We first define the critique structure, then outline the training method for critic.

\paragraph{Critique Structure}  
The critique generated by the critic model $\pi_{\phi}$ consists of two components:  \textit{Discrimination}, which involves analyzing and assessing the quality of the candidate action, and \textit{Revision}, which provides actionable suggestions for improvement.  

For the discrimination component, the critic model evaluates the candidate action in three predefined dimensions:  
\begin{itemize}[leftmargin=*]  
    \item \texttt{Contribution}: Assessing how well the candidate action contributes to solving the overall task. Irrelevant actions receive a low score.  
    \item \texttt{Feasibility}: Determining whether the candidate action is valid according to the agent's predefined action list.  
    \item \texttt{Efficiency}: Evaluating whether the candidate action approaches the task optimally, avoiding unnecessary steps or redundancy.  
\end{itemize}  

For the revision component, the critic model first assigns an overall grade to each candidate action (\ie \texttt{[Excellent/Good/Neutral/Poor/Very Poor]}), then generates concise and actionable suggestions based on its analysis of the candidate actions (see Appendix~\ref{sec:critic_prompt} for detailed prompts).  

\paragraph{Fine-tuning the Critic Model}  
To collect high-quality step-level expert critiques, we employ an expert critic annotator (e.g., GPT-4 \cite{openai2024gpt4technicalreport}) to generate critiques for each candidate action based on a reference expert trajectory.  
For a specific environment $e \in \mathcal{E}$, at time step $t$, the actor generates candidate actions $a_{t}$.  
Given the refined history trajectory $\tau_{t}^{\prime}$ and the expert trajectory $\tau^{\rm{exp}}$, 
the expert critic $\pi_{\rm{exp}}$ assesses the alignment between candidate actions and the optimal ones, converting this assessment into structured critiques according to a predefined format.  
Formally, the critiques for each candidate action are represented as:  
$
c_{t} \sim \pi_{\rm{exp}}(c \mid a_{t}, \tau_{t}^{\prime}, \tau^{\rm{exp}}, e).  
$ 

Subsequently, the actor refines its next step by generating \( a^{\prime}_{t} \) based on the critique \( c_{t} \).  
Critiques are collected only when \( \mathcal{R}(\tau^{\prime})\!=\!1 \), ensuring they originate from correct trajectories.  
We collect step-level expert critiques from each environment, forming the dataset  $\mathcal{D}_{\text {critique }}\!=\!\cup_{e\in \mathcal{E}} \mathcal{D}^{e}_{\text {critique }}$.  
The critic model is then fine-tuned using supervised learning with the collected expert critiques.  
Specifically, we apply the standard language modeling loss, defined as:  
\begin{equation*}  
    \mathcal{L}_{\text{critic}}(\phi) =  
    \mathbb{E}_{(c_{t}, \tau_{t}^{\prime}, a_{t}, e) \sim \mathcal{D}_{\text{critique}}} \Big[ \log \pi_\phi\left(c_{t} \mid \tau_{t}^{\prime}, a_{t}, e\right) \Big].  
\end{equation*}  

This approach enables the critic model to generate structured, step-level critiques that provide both discrimination and revision, which are crucial for guiding the actor model toward more effective decision-making.

\subsection{Action Refinement}
\label{sec:critic_utilization}
Although the critic model is trained to provide high-quality critiques, the actor may not fully utilize the feedback. Therefore, action refinement is necessary to enhance the actor’s ability to effectively leverage critiques for improving its actions.
A key challenge in this process is policy misalignment. 
During training, the actor model learns to incorporate critiques based on its current policy. 
However, after training, the model’s policy may evolve, making it difficult to integrate critiques effectively for newly generated candidate actions. 
This misalignment can lead to suboptimal action refinement.

To address this issue, we propose an iterative action refinement method based on supervised fine-tuning (SFT). 
This method consists of two main components: exploration and learning.  
In the exploration step at iteration \( k \), for a specific environment $e \in \mathcal{E}$, the actor model \( \pi_{\theta}^{k-1} \) interacts with the environment under the guidance of the critic model \( \pi_{\phi} \). At each time step \( t \), the refined action is given by \( a'_t = \pi_{\theta}(a \mid \tau_{t}^{\prime}, c_t, e) \), forming a set of critique-action pairs \( \{(\tau_{t}^{\prime}, c_t, a'_t)\}_{t=1}^{T} \). Here, \( \tau_{t}^{\prime} \) represents the previously refined trajectory, and \( c_t \) denotes the current critique. To ensure the quality of the trajectories, we filter them based on the environment reward \( \mathcal{R} \). Only trajectories for which \( \mathcal{R}(\tau') = 1 \), along with their corresponding critique-action pairs, are retained. 
We collect correct trajectories and critique-action pairs from each environment, yielding two datasets: \( \mathcal{D}_{\text{correct}}\!=\!\cup_{e \in \mathcal{E}} \mathcal{D}^{e}_{\text{correct}} \) and \( \mathcal{D}_{\text{refine}}\!=\! \cup_{e \in \mathcal{E}} \mathcal{D}^{e}_{\text{refine}} \). The dataset \( \mathcal{D}_{\text{correct}} \) enhances the model’s reasoning ability by providing correct trajectories, while \( \mathcal{D}_{\text{refine}} \) improves its capacity to utilize critiques and generate refined actions.

In the learning step of iteration \( k \), the newly collected datasets, combined with the expert path \( \mathcal{D}_{\text{expert}} \) for each environment, are used to fine-tune the actor model \( \pi_\theta \).  
To avoid overfitting, we follow previous work~\citep{zelikman2022starbootstrapping} by fine-tuning the original model \( \pi_\theta \) rather than the previous iteration model \( \pi_\theta^{k-1} \).  
Furthermore, following AgentTuning~\citep{zeng2023agenttuningenablinggeneralizedagent}, we incorporate general datasets such as ShareGPT3\footnote{\url{https://huggingface.co/datasets/Vtuber-plan/sharegpt-cleaned}} to improve generalization.
The training objective is defined as:
\[
\begin{aligned}
    \mathcal{L}_{\text{actor}}(\theta) &= \beta  \ \left\{ \mathbb{E}_{(\tau, x, e) \sim \mathcal{D}_{\text{train}}} \Big[\log \pi_\theta(\tau \mid x, e)\Big] + \mathbb{E}_{(a'_t,\tau'_{t}, c_t, e) \sim \mathcal{D}_{\text{refine}}} \Big[\log \pi_\theta(a'_t \mid \tau_{t}^{\prime}, c_t, e)\Big] \right\} \\
    &+ (1-\beta) \ \mathbb{E}_{(x, y) \sim \mathcal{D}_{\text{general}}} \Big[\log \pi_\theta(y \mid x)\Big],
\end{aligned}
\]
where \( \mathcal{D}_{\text{train}} = \mathcal{D}_{\text{expert}} \cup \mathcal{D}_{\text{correct}} \). 
After this refinement step, a new dataset with higher-quality samples is generated for further training.  
Overall, this iterative action refinement process allows the actor model to progressively improve its reasoning capabilities and better integrate critiques through continuous interaction with the environment.

\section{Experiment Settings}
In this section, we conduct extensive experiments in three interactive environments to demonstrate the effectiveness of the critic model and our \method framework.
\subsection{Interactive and Agentic Environments}
Following previous
work~\citep{xi2024agentgymevolvinglargelanguage, prasad-etal-2024-adapt}, we conduct experiments on three types of representative interactive environments:
\begin{itemize}[leftmargin=*]
    \item WebShop~\citep{yao2023webshop}, which is an interactive web environment for online shopping. It contains 12K instructions and offers over one million real products from amazon.com. Agents can click buttons on the webpage or perform searches using the search engine.
    \item ScienceWorld~\citep{wang2022scienceworlda}, which is a text-based scientific environment designed to evaluate agents' scientific reasoning abilities. It includes 30 types of scientific tasks at the standard elementary science curriculum level.
    \item TextCraft~\citep{prasad2024}, which is a text-based environment to create Minecraft items. 
    It constructs a crafting tree based on Minecraft's recipes. Each task provides a target item and a list of crafting commands generated by the tree. Agents receive a reward of 1 when they successfully craft the target item.
\end{itemize}
\paragraph{Evaluation Metrics}
Following the setup of AgentGym~\citep{xi2024agentgymevolvinglargelanguage}, we evaluate our model on the test sets for these three environments (200 simulations for ScienceWorld and WebShop, 100 for TextCraft). For ScienceWorld and WebShop, we use the average final score as the evaluation metric. 
For TextCraft, we use the success rate as the evaluation metric. 
Further details can be found in Appendix~\ref{sec:eval_settings}.

\subsection{Training Settings}
We use Llama-3-8B-Instruct~\citep{llama3modelcard} as the backbone model for both the actor and critic models.  
To collect training data, we randomly sample 500 simulations from WebShop, 350 from ScienceWorld, and 374 from TextCraft.  
We train the critic model by using the expert critic (\ie GPT-4o) to guide the actor in interacting with the environment three times, collecting expert critiques during each interaction.
For action refinement, we perform three iterations and report the results of the third iteration in Table~\ref{tab:main_results_2}. 
The training data for the critic model and each iteration of action refinement are provided in Table~\ref{tab:datasize}. Additional details can be found in Appendix~\ref{sec:implementation_details}.
\subsection{Baselines}  
To evaluate the effectiveness of our critic model, we fix the actor model as Llama-3-8B-Instruct~\citep{llama3modelcard}. We then compare our critic model against two types of approaches (see Appendix~\ref{sec:app_baselines} for implementation details):  
\begin{inparaenum}[\it 1)]  
\item \textbf{Numerical based}: We use DGAP~\citep{qian2025discriminatorguided}, a discriminator trained to assess the alignment between actor actions and expert actions at the step level, and Explicit RM~\citep{xia2025agentrmenhancingagentgeneralization}, which is trained to predict the Q-value, \ie expected accumulated rewards at each time step.
\item \textbf{Verbal based}: We employ a self-critique method where the actor model itself  generates critiques for each candidate action at the step level. Additionally, we use GPT-4o (gpt-4o-2024-08-06)~\citep{openai2024gpt4technicalreport} as a critic, which serves as a strong general-purpose evaluator. The structure of the critiques aligns with that in Appendix~\ref{sec:critic_prompt}.
\end{inparaenum}
Each approach guides the actor's inference, and we evaluate their effectiveness based on the actor's performance\footnote{In the subsequent experiments of this paper, we set the number of candidate actions sampled at inference time to $M=5$.}.

For the assessment of \method, consistent with the methodology used in AgentGym~\citep{xi2024agentgymevolvinglargelanguage}, we select a combination of closed-source models, including GPT-3.5-turbo~\citep{gpt3.5}, GPT-4o~\citep{gpt4}, Claude 3~\citep{anthropic2023claude}, and DeepSeek-Chat~\citep{deepseekai2025deepseekv3technicalreport}, as well as open-source models such as Llama-3-70B-Instruct~\citep{llama3modelcard}. 
Additionally, we include agents trained on expert trajectories, such as AgentLM (13B and 70B)~\citep{zeng-etal-2024-agenttuning} and Agent-Flan~\citep{chen-etal-2024-agent}. 
We also compare our approach to Iterative SFT, which iteratively refines the actor model using correct trajectories collected from interactions with the environment, and Reflexion~\citep{shinn2023reflexion}, a self-reinforcement method that concludes each iteration with a summary to guide decision-making in subsequent iterations. We report the results from the third iteration for both approaches.

\section{Main Results}
\label{sec:main_results}
\setlength\tabcolsep{13pt}
\begin{table}[t]
  \centering
  \caption{We compared the natural language feedback provided by the trained critic model with other methods, including numerical-based (\ie DGAP, Explict RM), and verbal-based (\ie self-critique, GPT-4o). Our critic offers better guidance to the actor model in all three interactive scenarios.}
\vspace{0.2cm}
\scalebox{0.75}{
    \begin{tabular}{llcccc}
    \toprule
    \textbf{Model} & \textbf{Method} & \textbf{WebShop} & \textbf{ScienceWorld} & \textbf{TextCraft} & \textbf{Average}\\
    \midrule
    \multirow{6}{*}{\textit{Llama-3-8B-Instruct}} 
    &\cellcolor{gray!15}No Critique &\cellcolor{gray!15}13.49 &\cellcolor{gray!15}14.48 &\cellcolor{gray!15}10.00 &\cellcolor{gray!15}12.65 \\
    & DGAP  &30.41 & 19.52   &21.00  &23.64 \\
    & Explicit RM &14.21 & 18.41   &15.00  &15.87 \\
    & Self-Critique  &1.50 & 10.06   &19.00 &10.19 \\
    & GPT-4o & 17.78 & 33.06 & 46.00 & 32.28 \\
    &\cellcolor{lightblue!100}Critic Model (\textit{Ours})  &\cellcolor{lightblue!100}\textbf{56.80} &\cellcolor{lightblue!100}\textbf{68.51}   &\cellcolor{lightblue!100}\textbf{59.00} &\cellcolor{lightblue!100}\textbf{61.44} \\ 
    \midrule
    \multirow{6}{*}{\textit{Llama-3-70B-Instruct}} 
    &\cellcolor{gray!15}No Critique &\cellcolor{gray!15}8.35 &\cellcolor{gray!15}49.20 &\cellcolor{gray!15}2.00 &\cellcolor{gray!15}19.85 \\
    & DGAP  &11.17  &55.86    &9.00  &25.34  \\ 
    & Explicit RM &11.18 & 48.41   &14.00  &24.53 \\
    & Self-Critique  &3.00 & 28.17   &40.00  &23.72 \\ 
    & GPT-4o &16.24  &43.73   &56.00  &38.65  \\
    &\cellcolor{lightblue!100}Critic Model (\textit{Ours}) &\cellcolor{lightblue!100}\textbf{52.20}   &\cellcolor{lightblue!100}\textbf{72.44}    &\cellcolor{lightblue!100}\textbf{73.00}  &\cellcolor{lightblue!100}\textbf{65.88}  \\ 
    \midrule
    \multirow{6}{*}{%
      \textit{Llama-3-8B-Instruct + SFT}
    }
    &\cellcolor{gray!15}No Critique  &\cellcolor{gray!15}\textbf{76.12} &\cellcolor{gray!15}32.85 &\cellcolor{gray!15}46.00 &\cellcolor{gray!15}51.66 \\ 
    & DGAP  &73.97 & 38.26   & 56.00 &56.08 \\
    & Explicit RM &74.33 & 41.57   &52.00  &55.97 \\
    & Self-Critique &50.18 & 31.22   &21.00 &34.13 \\
    & GPT-4o & 55.65 & 48.48  &\textbf{58.00}  & 54.04 \\
    &\cellcolor{lightblue!100}Critic Model (\textit{Ours}) &\cellcolor{lightblue!100}\underline{74.68}  &\cellcolor{lightblue!100}\textbf{55.94}  &\cellcolor{lightblue!100}\underline{56.00} &\cellcolor{lightblue!100}\textbf{62.21} \\ 
    \bottomrule
    \end{tabular}
    }
  \label{tab:critic_results}
  \vspace{-10pt}
\end{table}

We first evaluate our critic model against numerical- and verbal-based methods to assess its effectiveness.  
As shown in Table~\ref{tab:critic_results}, our 8B critic model significantly outperforms GPT-4o on both the Llama-3 (8B and 70B) and Llama-3-8B's fine-tuned variant trained on expert data.  
Furthermore, Table~\ref{tab:main_results_2} demonstrates that through iterative action refinement of the actor model, our \method notably enhances actor performance in interactive environments.  
It outperforms both advanced closed-source models (e.g., GPT-4o) and agents trained on expert trajectories (e.g., AgentLM-70B and Agent-FLAN).
Based on these results, we identify three key findings. 

\paragraph{Finding 1: Verbal critique feedback is more effective than numerical signal.} 
As shown in Table~\ref{tab:critic_results}, the Critic Model consistently outperforms numerical feedback from the discriminator. For the Llama-3-8B model, it achieves an average improvement of 42.89\% over the no-critique baseline, while the discriminator yields only a 5.09\% gain. 
The advantage persists with the stronger Llama-3-70B and the fine-tuned Llama-3-8B models, where the Critic Model delivers a 46.03\% and 10.55\% improvement, respectively, compared to 5.49\% and 4.42\% from the discriminator.
These results indicate that the discriminator's numerical feedback relies heavily on the model's inherent capabilities, as it essentially performs action-level filtering. 
For example, in the ScienceWorld, the discriminator enables Llama-3-70B to reach 55.86\%, but only 18.52\% for Llama-3-8B. 
In contrast, the Critic Model shows minimal sensitivity to the base model, achieving 68.51\% with Llama-3-8B and 72.44\% with Llama-3-70B.
However, the self-critique approach, which also relies on verbal feedback, shows a negative effect on both Llama-3-70B, 8B and its fine-tuned variant. This suggests that while verbal feedback can convey richer information, untrained self-critique methods often produce low-quality or unstructured feedback, failing to provide effective guidance.
\vspace{-5pt}

\begin{table*}[t!]
\setlength\tabcolsep{25pt}
\centering
\caption{Results from three interactive environments. The \method here employs a two-player setting with a fine-tuned LLama-3-8B critic model and an 8B actor model refined through three iterations. We compare its performance with various models.}
\label{tab:main_results_2}
\scalebox{0.77}{
\begin{tabular}{lcccccccc}
\toprule
\textbf{Method} & \textbf{WebShop} & \textbf{ScienceWorld} & \textbf{TextCraft} & \textbf{Average} \\
\midrule
\rowcolor[gray]{0.95}\multicolumn{5}{c}{\textit{Closed-source Models}} \\
DeepSeek-Chat &11.00 &16.80 &23.00 &16.93 \\
Claude-3-Haiku &5.50 &0.83 &0.00 &2.11 \\
Claude-3-Sonnet &1.50 &2.78 &38.00 &14.09 \\
GPT-3.5-Turbo &12.50 &7.64 &47.00 &22.38 \\
GPT-4o & 25.48 & 46.91 & 64.00 & 45.46 \\
\midrule
\rowcolor[gray]{0.95}\multicolumn{5}{c}{\textit{Open-source Models}} \\
Llama-3-70B-Instruct & 8.35 & 49.20 & 2.00 & 19.85 \\
AgentLM-13B & 39.50 & 2.75 & 0.00 & 14.08 \\
AgentLM-70B & 49.50 & 10.68 & 4.00 & 21.39 \\
Agent-FLAN & 40.35 & 28.64 & 16.00 & 28.33 \\
\midrule
Llama-3-8B-Instruct & 13.49 & 14.48 & 10.00 & 12.66 \\
\ w/\ Reflexion &14.08 &12.55 &8.00 &11.54 \\
\ w/\ Iterative SFT & \textbf{78.21}	&41.42 	&55.00 	&58.21  \\
\rowcolor{lightblue!100} \ w/\ \textbf{\method(\textit{Ours})} &\underline{76.17} &\textbf{78.43} &\textbf{68.00} &\textbf{74.20} \\
\bottomrule
\end{tabular}
}
\end{table*}

\begin{figure*}[t]
    \centering
       \vspace{-0.1cm}    \includegraphics[width=0.92\linewidth]{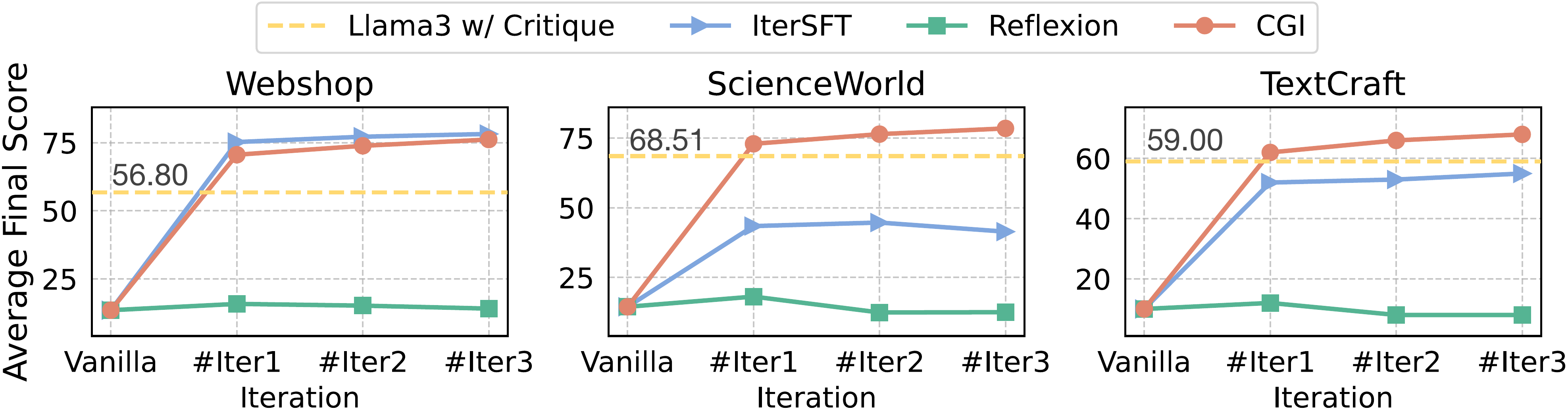}
    \caption{Performance of different iterative methods across iterations in three agentic tasks.}
    \label{fig:main_02}
    \vspace{-0.4cm}
\end{figure*}

\paragraph{Finding 2: Fine-tuned models struggle to utilize critiques.}
While supervised fine-tuning significantly improves baseline performance (+33.11\%), our experiments reveal a key limitation: fine-tuned models struggle to effectively incorporate critique feedback. 
In the ScienceWorld scenario, Vanilla Llama3-8B and Llama3-70B show substantial improvements with Critic guidance, achieving average scores of 68.51\% and 72.44\%, respectively. 
However, despite having a stronger base performance, fine-tuned Llama3-8B only reaches 55.94\% with the same critique mechanism. 
This issue persists in the WebShop scenario, where critique guidance even causes a performance decline in fine-tuned Llama3-8B, dropping from 76.12\% to 74.68\%.  
These results suggest that while fine-tuning enhances base performance, it may reduce the model's responsiveness to external feedback. For example, in the ScienceWorld scenario, despite the critic model advising the actor to "look around" and assess the environment for possible actions, the actor persists in executing ``Go to Bathroom'', an action that seems closest to the task description. 
This behavior disregards the fact that the actor cannot reach the bathroom from its current position, leading to a deadlock (see Appendix~\ref{app:sft_sample}).  
Our iterative action refinement approach helps mitigate this issue; after three iterations, the model shows a significant improvement in critique utilization, with its performance surpassing that of the actor model with only supervised fine-tuning by +15.99\%, achieving state-of-the-art results compared to various baselines.

\paragraph{Finding 3: \method continuously enhance model performance via action refinement.}
As shown in Figure~\ref{fig:main_02}, compared to other iterative methods (\eg Reflexion and vanilla iterative SFT), \method consistently supports model performance improvement. 
In contrast, Reflexion shows minimal improvement and can even cause performance degradation. This aligns with previous findings~\citep{shinn2023reflexion}, where self-critique struggles to escape local minima in agentic tasks that require significant diversity and exploration. 
Iterative SFT achieves a notable improvement only in the first iteration, with minimal progress in the following 2-3 iterations. 
This issue is more pronounced in long-horizon tasks (\eg ScienceWorld and TextCraft), suggesting that the model may over-sample simpler tasks while under-sampling more complex ones. Consequently, the training set for subsequent iterations becomes dominated by easier tasks, with few solutions for harder ones. 
As iterations progress, this bias deepens, leading to a long-tail distribution where solutions to more challenging tasks are scarce, ultimately causing the model to plateau or even degrade. 
In contrast, \method leverages high-quality critiques and enhances the actor's utilization, enabling the model to consistently sample solutions to more difficult tasks, thus supporting sustained performance improvement.

\section{Further Analysis}

\begin{figure*}[t]
    \centering
    \includegraphics[width=0.9\linewidth]{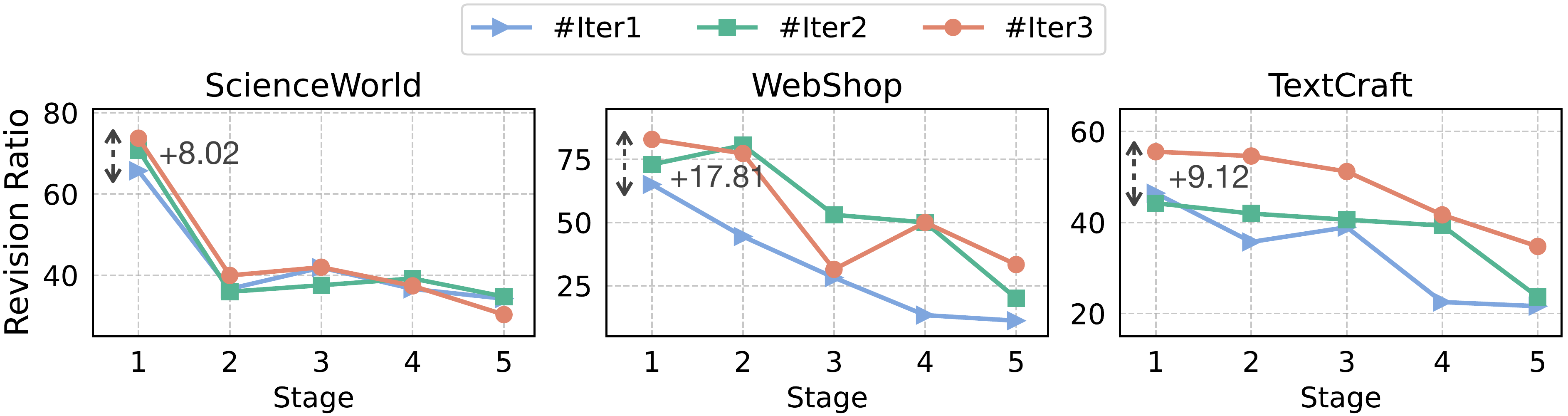}
    \caption{Revision Ration of actor model at different trajectory stages across three tasks. \#Iter-$k$ denotes the $k$-th iteration of action refinement. Revision Ratio represents the proportion of actions that differ after critique compared to those without critique.}
    \label{fig:analysis_02}
\end{figure*}

\subsection{Qualitative Analysis of \method}  
In this section, we conduct a qualitative analysis to examine how \method enhances performance in agentic tasks.  
Specifically, we focus on two key aspects:  
\begin{inparaenum}[\it 1)]
\item \textbf{Trajectory-level}: Identifying the stages in the trajectory where \method provides the most significant improvement;
\item \textbf{Task-level}: Evaluating \method's performance across tasks of varying trajectory length (see Appendix~\ref{sec:eval_settings} for categorization).  
\end{inparaenum}  

\paragraph{\method significantly improves early-stage performance.}  
To determine at which stage of the trajectory \method contributes most to performance improvements, we divide the trajectory into five stages based on its length and compute the extent to which the critique influences the actor model's behavior, measured by the Revision Ratio.  
As shown in Figure~\ref{fig:analysis_02}, across three tasks, actor models exhibit the highest action revision frequency in stage 1, with the revision ratio dropping sharply in later stages.  
This suggests that the critique primarily guides the actor during early exploration, helping reduce ineffective searches.  
Notably, as the actor undergoes more refinement iterations, its revision ratio in stage 1 increases (+8.02\% on ScienceWorld, +17.81\% on WebShop, +9.12\% on TextCraft), indicating that better critique utilization accelerates effective exploration.  
This, in turn, enhances \method's efficiency, enabling it to achieve higher scores in fewer steps (Figure~\ref{fig:app_02}).

\begin{wrapfigure}{r}{0.43\textwidth}
    \centering
    \vspace{-0.8cm}
    \includegraphics[width=\linewidth]{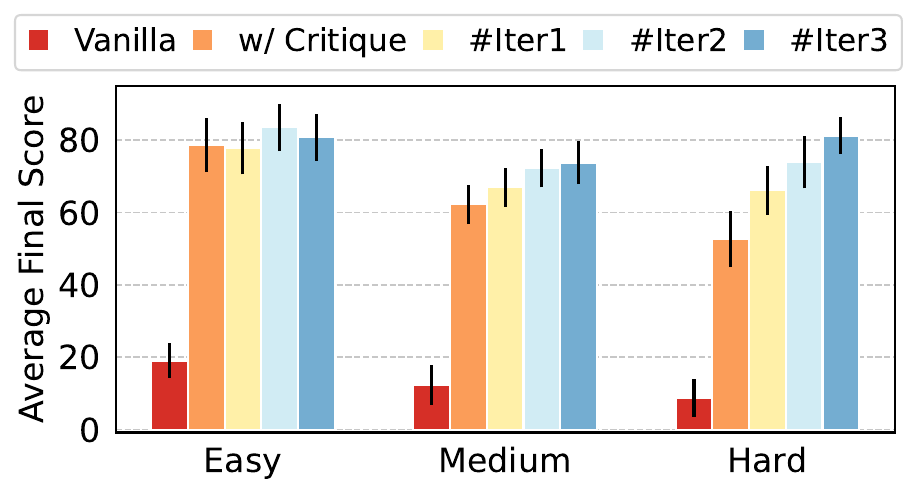}
    \caption{Performance of models across varying trajectory length. Vanilla refers to the Llama3-8B model, w/ Critique refers to the critique-guided inference with our trained Critic-Llama3, and \#Iter-$k$ represents the $k$-th iteration of action refinement.}
    \label{fig:analysis_01}
    \vspace{-8pt} 
\end{wrapfigure}

\paragraph{\method helps the model to continuously improve on long-horizon tasks.}

To investigate the effect of \method on tasks of varying trajectory length, we categorize the ScienceWorld scenarios into three difficult-level groups based on the average length of the oracle agent's trajectories. 
Longer trajectories correspond to higher difficulty levels. 
As shown in Figure~\ref{fig:analysis_01}, the vanilla model (Llama3-8B) performs poorly, with performance decreasing as the task length increases. 
Critique-Guided Inference leads to a significant improvement, particularly on easy tasks, where performance increases by +59.64\%. Although there is also a notable improvement on harder tasks (+43.95\%), the gain is less pronounced compared to easier tasks. 
However, with iterative action refinement, the model exhibits the greatest improvement on hard tasks, with a +28.75\% increase after three iterations.
These results suggest that models without action refinement can quickly adapt to critiques for easy and medium-level tasks. However, for longer tasks, the model’s ability to leverage critiques diminishes. 
Action refinement helps the model continuously follow critiques, thereby enhancing performance on long-horizon tasks.

\subsection{Effects of Number of Candidate Actions}
\begin{wrapfigure}{r}{0.42\textwidth}
\vspace{-0.25cm} 
    \centering
    \includegraphics[width=0.95\linewidth]{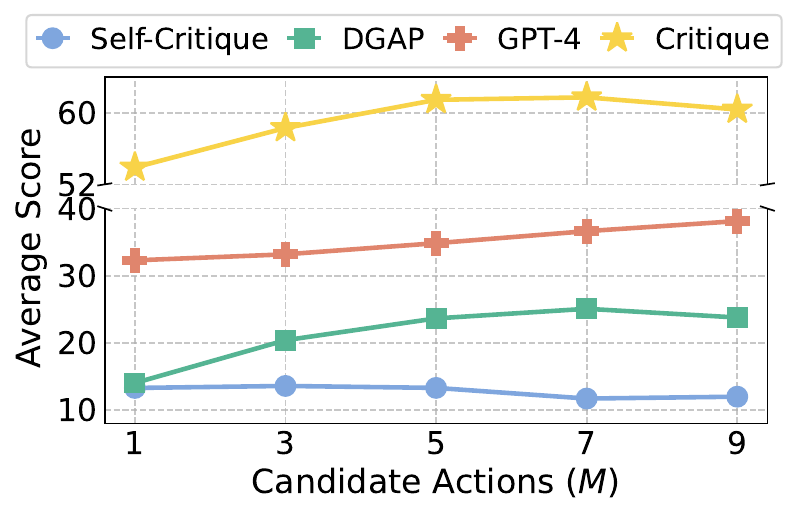}
    \caption{Performance of different methods across varying candidate actions. The average score is the mean final score across three environments.}
    \vspace{-0.3cm} 
    \label{fig:scalability}
\end{wrapfigure}
To further evaluate the scalability of \method, we varied the number of candidate actions ($M$) sampled at each step during inference. 
As shown in Figure~\ref{fig:scalability}, we observe the following trends:
\method achieves a significant improvement with just one candidate action, reaching an average score of 56.89\%. 
As $M$ increases, performance improves further, rising from 56.89\% to 61.72\%. 
However, when $M=7$, performance plateaus, indicating saturation.
Other numerical- and verbal-based methods exhibit similar trends. 
For GPT-4 as a critic, performance increases with $M$, improving from 32.28\% to 38.13\%. DGAP is particularly sensitive to the number of candidate actions, with performance rising from 15.02\% to 25.07\%. 
This highlights that numerical methods are highly dependent on the quality of the sampled candidate actions, as they only discriminate but not suggest meaningful revisions.
In contrast, self-critique does not benefit from an increase in $M$. This suggests that low-quality critiques can degrade performance, regardless of the number of candidate actions.

\section{Ablation Study}
\begin{wrapfigure}{r}{0.42\textwidth}
    \centering
    \vspace{-0.4cm}
    \includegraphics[width=\linewidth]{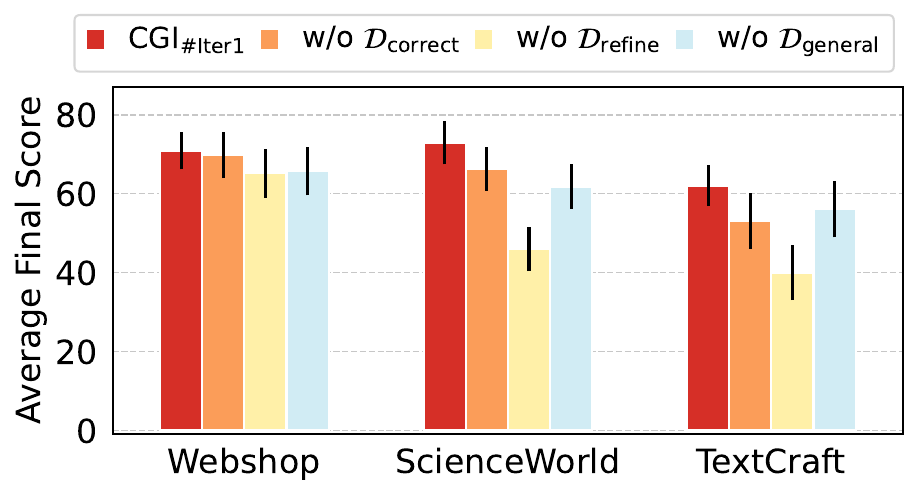}
    \caption{Ablation study of the three data types used in action refinement: $\mathcal{D}_{\text{correct}}$ (correct trajectories to improve the actor's reasoning ability), $\mathcal{D}_{\text{refine}}$ (critique-action pairs to enhance critique utilization), and $\mathcal{D}_{\text{general}}$ (data to maintain the model's generalization capability). \method${_{\#\text{Iter1}}}$ refers to the actor model trained in the first iteration of action refinement using the full dataset.}
    \label{fig:ablation}
    \vspace{-8pt}
\end{wrapfigure}

To analyze the contribution of the three types of data (\ie $\mathcal{D}_{\text{correct}}$, $\mathcal{D}_{\text{refine}}$, and $\mathcal{D}_{\text{general}}$) used to enhance the actor model during the action refinement stage, we conduct an ablation study. 
Specifically, we remove each dataset from the training data used in the first iteration of action refinement and evaluate the final performance under the guidance of the critic model.  
As shown in Figure~\ref{fig:ablation}, removing any one of the three datasets results in a performance drop compared to the actor model trained with the full dataset (\method$_{\#\text{Iter1}}$). 
The most significant decrease occurs when the critique-action pairs are removed (w/o $\mathcal{D}_{\text{refine}}$), where the average performance across the three tasks drops from 68.50\% to 50.37\%. 
This decline is more pronounced for longer tasks, with performance reductions of 5.4\% on WebShop, 22\% on TextCraft, and 26.94\% on ScienceWorld. 
These results highlight that enhancing the actor's critique utilization ability is the most critical factor in improving overall performance.
A less severe drop is observed when the generalization data set is excluded (w/o $\mathcal{D}_{\text{general}}$), highlighting the importance of general instructions for model generalization, which helps the model adapt to the unseen test set.

\section{Conclusion}
We have introduced \textbf{C}ritique-\textbf{G}uided \textbf{I}mprovement (\method), a two-player framework that emphasizes nature language feedback for iterative refinement of LLM-based agents. 
By separating the roles of an actor, which proposes actions, and a critic, which provides verbal guidance, \method circumvents the limitations of purely numerical signals and addresses the challenges arising from self-refinement. 
Experimental results in three interactive and agentic environments confirm the effectiveness of \method, with a small critic model outperforming GPT-4 in providing feedback. 
Our findings highlight that explicit verbal-based critiques enable more flexible action revision and deeper exploration, ultimately driving significant improvements in state-of-the-art performance. 
Future work may investigate additional domains, more advanced training procedures for the critic, and new ways of integrating critiques into the agent's action-planning mechanisms.


\section*{Acknowledgement}
We appreciate the support from the Chinese NSF General Program (No.62572129), Major Research Plan (No.92270121).
We also acknowledge the use of an icon from Flaticon\footnote{\url{https://www.flaticon.com}} and thank its creators for providing this visually appealing design.

\bibliography{custom,anthology}

\begin{thebibliography}{10}

\bibitem{openai2024gpt4technicalreport}
OpenAI.
\newblock {GPT-4 Technical Report}, 2024.

\bibitem{Gemini2023}
{Gemini Team}.
\newblock Gemini: A family of highly capable multimodal models.
\newblock Technical report, Google, 12 2023.

\bibitem{llama3modelcard}
Meta.
\newblock Llama 3 model card.
\newblock 2024.

\bibitem{durante2024agentaisurveyinghorizons}
Zane Durante, Qiuyuan Huang, Naoki Wake, Ran Gong, Jae~Sung Park, Bidipta Sarkar, Rohan Taori, Yusuke Noda, Demetri Terzopoulos, Yejin Choi, Katsushi Ikeuchi, Hoi Vo, Li~Fei-Fei, and Jianfeng Gao.
\newblock {Agent AI: Surveying the Horizons of Multimodal Interaction}, 2024.

\bibitem{chen2024personapersonalizationsurveyroleplaying}
Jiangjie Chen, Xintao Wang, Rui Xu, Siyu Yuan, Yikai Zhang, Wei Shi, Jian Xie, Shuang Li, Ruihan Yang, Tinghui Zhu, Aili Chen, Nianqi Li, Lida Chen, Caiyu Hu, Siye Wu, Scott Ren, Ziquan Fu, and Yanghua Xiao.
\newblock {From Persona to Personalization: A Survey on Role-Playing Language Agents}, 2024.

\bibitem{li2024selu}
Boyu Li, Haobin Jiang, Ziluo Ding, Xinrun Xu, Haoran Li, Dongbin Zhao, and Zongqing Lu.
\newblock {Selu: Self-learning embodied mllms in unknown environments}.
\newblock {\em arXiv preprint arXiv:2410.03303}, 2024.

\bibitem{yuan2024easytoolenhancingllmbasedagents}
Siyu Yuan, Kaitao Song, Jiangjie Chen, Xu~Tan, Yongliang Shen, Ren Kan, Dongsheng Li, and Deqing Yang.
\newblock {EASYTOOL: Enhancing LLM-based Agents with Concise Tool Instruction}, 2024.

\bibitem{qin2023toolllmfacilitatinglargelanguage}
Yujia Qin, Shihao Liang, Yining Ye, Kunlun Zhu, Lan Yan, Yaxi Lu, Yankai Lin, Xin Cong, Xiangru Tang, Bill Qian, Sihan Zhao, Lauren Hong, Runchu Tian, Ruobing Xie, Jie Zhou, Mark Gerstein, Dahai Li, Zhiyuan Liu, and Maosong Sun.
\newblock {ToolLLM: Facilitating Large Language Models to Master 16000+ Real-world APIs}, 2023.

\bibitem{hu2025selfevolving}
Yue Hu, Yuzhu Cai, Yaxin Du, Xinyu Zhu, Xiangrui Liu, Zijie Yu, Yuchen Hou, Shuo Tang, and Siheng Chen.
\newblock {Self-Evolving Multi-Agent Collaboration Networks for Software Development}.
\newblock In {\em The Thirteenth International Conference on Learning Representations}, 2025.

\bibitem{jimenez2024swebench}
Carlos~E Jimenez, John Yang, Alexander Wettig, Shunyu Yao, Kexin Pei, Ofir Press, and Karthik~R Narasimhan.
\newblock {SWE}-bench: Can language models resolve real-world github issues?
\newblock In {\em The Twelfth International Conference on Learning Representations}, 2024.

\bibitem{yao2023webshop}
Shunyu Yao, Howard Chen, John Yang, and Karthik Narasimhan.
\newblock {WebShop: Towards Scalable Real-World Web Interaction with Grounded Language Agents}, 2023.

\bibitem{putta2024agentqadvancedreasoning}
Pranav Putta, Edmund Mills, Naman Garg, Sumeet Motwani, Chelsea Finn, Divyansh Garg, and Rafael Rafailov.
\newblock {Agent Q}: Advanced reasoning and learning for autonomous ai agents, 2024.

\bibitem{wang-etal-2024-math}
Peiyi Wang, Lei Li, Zhihong Shao, Runxin Xu, Damai Dai, Yifei Li, Deli Chen, Yu~Wu, and Zhifang Sui.
\newblock Math-shepherd: Verify and reinforce {LLM}s step-by-step without human annotations.
\newblock In Lun-Wei Ku, Andre Martins, and Vivek Srikumar, editors, {\em Proceedings of the 62nd Annual Meeting of the Association for Computational Linguistics (Volume 1: Long Papers)}, pages 9426--9439, Bangkok, Thailand, August 2024. Association for Computational Linguistics.

\bibitem{qian2025discriminatorguided}
Haofu Qian, Chenjia Bai, Jiatao Zhang, Fei Wu, Wei Song, and Xuelong Li.
\newblock {Discriminator-Guided Embodied Planning for {LLM} Agent}.
\newblock In {\em The Thirteenth International Conference on Learning Representations}, 2025.

\bibitem{xia2025agentrmenhancingagentgeneralization}
Yu~Xia, Jingru Fan, Weize Chen, Siyu Yan, Xin Cong, Zhong Zhang, Yaxi Lu, Yankai Lin, Zhiyuan Liu, and Maosong Sun.
\newblock {AgentRM}: Enhancing agent generalization with reward modeling, 2025.

\bibitem{chen2025autonomous}
Zhenfang Chen, Delin Chen, Rui Sun, Wenjun Liu, and Chuang Gan.
\newblock Autonomous agents from automatic reward modeling and planning.
\newblock In {\em The Thirteenth International Conference on Learning Representations}, 2025.

\bibitem{stiennon2022learningsummarizehumanfeedback}
Nisan Stiennon, Long Ouyang, Jeff Wu, Daniel~M. Ziegler, Ryan Lowe, Chelsea Voss, Alec Radford, Dario Amodei, and Paul Christiano.
\newblock Learning to summarize from human feedback, 2022.

\bibitem{ouyang2022training}
Long Ouyang, Jeffrey Wu, Xu~Jiang, Diogo Almeida, Carroll Wainwright, Pamela Mishkin, Chong Zhang, Sandhini Agarwal, Katarina Slama, Alex Gray, John Schulman, Jacob Hilton, Fraser Kelton, Luke Miller, Maddie Simens, Amanda Askell, Peter Welinder, Paul Christiano, Jan Leike, and Ryan Lowe.
\newblock {Training language models to follow instructions with human feedback}.
\newblock In Alice~H. Oh, Alekh Agarwal, Danielle Belgrave, and Kyunghyun Cho, editors, {\em Advances in Neural Information Processing Systems}, 2022.

\bibitem{madaan2024self}
Aman Madaan, Niket Tandon, Prakhar Gupta, Skyler Hallinan, Luyu Gao, Sarah Wiegreffe, Uri Alon, Nouha Dziri, Shrimai Prabhumoye, Yiming Yang, et~al.
\newblock {Self-refine: Iterative refinement with self-feedback}.
\newblock {\em Advances in Neural Information Processing Systems}, 36, 2024.

\bibitem{shinn2023reflexion}
Noah Shinn, Federico Cassano, Edward Berman, Ashwin Gopinath, Karthik Narasimhan, and Shunyu Yao.
\newblock {Reflexion: Language Agents with Verbal Reinforcement Learning}, 2023.

\bibitem{yang2024selfgoal}
Ruihan Yang, Jiangjie Chen, Yikai Zhang, Siyu Yuan, Aili Chen, Kyle Richardson, Yanghua Xiao, and Deqing Yang.
\newblock {SELFGOAL: Your Language Agents Already Know How to Achieve High-level Goals}.
\newblock In {\em NeurIPS 2024 Workshop on Open-World Agents}, 2024.

\bibitem{kumar2025training}
Aviral Kumar, Vincent Zhuang, Rishabh Agarwal, Yi~Su, John~D Co-Reyes, Avi Singh, Kate Baumli, Shariq Iqbal, Colton Bishop, Rebecca Roelofs, Lei~M Zhang, Kay McKinney, Disha Shrivastava, Cosmin Paduraru, George Tucker, Doina Precup, Feryal Behbahani, and Aleksandra Faust.
\newblock {Training Language Models to Self-Correct via Reinforcement Learning}.
\newblock In {\em The Thirteenth International Conference on Learning Representations}, 2025.

\bibitem{cuadron2025dangeroverthinkingexaminingreasoningaction}
Alejandro Cuadron, Dacheng Li, Wenjie Ma, Xingyao Wang, Yichuan Wang, Siyuan Zhuang, Shu Liu, Luis~Gaspar Schroeder, Tian Xia, Huanzhi Mao, Nicholas Thumiger, Aditya Desai, Ion Stoica, Ana Klimovic, Graham Neubig, and Joseph~E. Gonzalez.
\newblock The danger of overthinking: Examining the reasoning-action dilemma in agentic tasks, 2025.

\bibitem{metz2025reward}
Yannick Metz, Andras Geiszl, Rapha{\"e}l Baur, and Mennatallah El-Assady.
\newblock {Reward Learning from Multiple Feedback Types}.
\newblock In {\em The Thirteenth International Conference on Learning Representations}, 2025.

\bibitem{wang2022scienceworlda}
Ruoyao Wang, Peter Jansen, Marc-Alexandre Côté, and Prithviraj Ammanabrolu.
\newblock {ScienceWorld: Is your Agent Smarter than a 5th Grader?}, 2022.

\bibitem{prasad2024}
Archiki Prasad, Alexander Koller, Mareike Hartmann, Peter Clark, Ashish Sabharwal, Mohit Bansal, and Tushar Khot.
\newblock {ADaPT: As-Needed Decomposition and Planning with Language Models}, 2024.

\bibitem{wang-etal-2024-reward-modeling}
Binghai Wang, Rui Zheng, Lu~Chen, Zhiheng Xi, Wei Shen, Yuhao Zhou, Dong Yan, Tao Gui, Qi~Zhang, and Xuanjing Huang.
\newblock Reward modeling requires automatic adjustment based on data quality.
\newblock In Yaser Al-Onaizan, Mohit Bansal, and Yun-Nung Chen, editors, {\em Findings of the Association for Computational Linguistics: EMNLP 2024}, pages 4041--4064, Miami, Florida, USA, November 2024. Association for Computational Linguistics.

\bibitem{cobbe2021trainingverifierssolvemath}
Karl Cobbe, Vineet Kosaraju, Mohammad Bavarian, Mark Chen, Heewoo Jun, Lukasz Kaiser, Matthias Plappert, Jerry Tworek, Jacob Hilton, Reiichiro Nakano, Christopher Hesse, and John Schulman.
\newblock {Training Verifiers to Solve Math Word Problems}, 2021.

\bibitem{yu-etal-2024-ovm}
Fei Yu, Anningzhe Gao, and Benyou Wang.
\newblock {OVM}, outcome-supervised value models for planning in mathematical reasoning.
\newblock In Kevin Duh, Helena Gomez, and Steven Bethard, editors, {\em Findings of the Association for Computational Linguistics: NAACL 2024}, pages 858--875, Mexico City, Mexico, June 2024. Association for Computational Linguistics.

\bibitem{luo2024improvemathematicalreasoninglanguage}
Liangchen Luo, Yinxiao Liu, Rosanne Liu, Samrat Phatale, Meiqi Guo, Harsh Lara, Yunxuan Li, Lei Shu, Yun Zhu, Lei Meng, Jiao Sun, and Abhinav Rastogi.
\newblock {Improve Mathematical Reasoning in Language Models by Automated Process Supervision}, 2024.

\bibitem{nakano2022webgptbrowserassistedquestionansweringhuman}
Reiichiro Nakano, Jacob Hilton, Suchir Balaji, Jeff Wu, Long Ouyang, Christina Kim, Christopher Hesse, Shantanu Jain, Vineet Kosaraju, William Saunders, Xu~Jiang, Karl Cobbe, Tyna Eloundou, Gretchen Krueger, Kevin Button, Matthew Knight, Benjamin Chess, and John Schulman.
\newblock {WebGPT: Browser-assisted question-answering with human feedback}, 2022.

\bibitem{bai2022constitutionalaiharmlessnessai}
Yuntao Bai, Saurav Kadavath, Sandipan Kundu, Amanda Askell, Jackson Kernion, Andy Jones, Anna Chen, Anna Goldie, Azalia Mirhoseini, Cameron McKinnon, Carol Chen, Catherine Olsson, Christopher Olah, Danny Hernandez, Dawn Drain, Deep Ganguli, Dustin Li, Eli Tran-Johnson, Ethan Perez, Jamie Kerr, Jared Mueller, Jeffrey Ladish, Joshua Landau, Kamal Ndousse, Kamile Lukosuite, Liane Lovitt, Michael Sellitto, Nelson Elhage, Nicholas Schiefer, Noemi Mercado, Nova DasSarma, Robert Lasenby, Robin Larson, Sam Ringer, Scott Johnston, Shauna Kravec, Sheer~El Showk, Stanislav Fort, Tamera Lanham, Timothy Telleen-Lawton, Tom Conerly, Tom Henighan, Tristan Hume, Samuel~R. Bowman, Zac Hatfield-Dodds, Ben Mann, Dario Amodei, Nicholas Joseph, Sam McCandlish, Tom Brown, and Jared Kaplan.
\newblock {Constitutional AI: Harmlessness from AI Feedback}, 2022.

\bibitem{zheng2023judgingllmasajudgemtbenchchatbot}
Lianmin Zheng, Wei-Lin Chiang, Ying Sheng, Siyuan Zhuang, Zhanghao Wu, Yonghao Zhuang, Zi~Lin, Zhuohan Li, Dacheng Li, Eric~P. Xing, Hao Zhang, Joseph~E. Gonzalez, and Ion Stoica.
\newblock {Judging LLM-as-a-Judge with MT-Bench and Chatbot Arena}, 2023.

\bibitem{madaan2023selfrefine}
Aman Madaan, Niket Tandon, Prakhar Gupta, Skyler Hallinan, Luyu Gao, Sarah Wiegreffe, Uri Alon, Nouha Dziri, Shrimai Prabhumoye, Yiming Yang, Shashank Gupta, Bodhisattwa~Prasad Majumder, Katherine Hermann, Sean Welleck, Amir Yazdanbakhsh, and Peter Clark.
\newblock {Self-Refine: Iterative Refinement with Self-Feedback}, 2023.

\bibitem{yao2023react}
Shunyu Yao, Jeffrey Zhao, Dian Yu, Nan Du, Izhak Shafran, Karthik~R Narasimhan, and Yuan Cao.
\newblock {ReAct: Synergizing Reasoning and Acting in Language Models}.
\newblock In {\em The Eleventh International Conference on Learning Representations}, 2023.

\bibitem{shang2025agentsquareautomaticllmagent}
Yu~Shang, Yu~Li, Keyu Zhao, Likai Ma, Jiahe Liu, Fengli Xu, and Yong Li.
\newblock {AgentSquare: Automatic LLM Agent Search in Modular Design Space}, 2025.

\bibitem{chen2024automanualconstructinginstructionmanuals}
Minghao Chen, Yihang Li, Yanting Yang, Shiyu Yu, Binbin Lin, and Xiaofei He.
\newblock {AutoManual: Constructing Instruction Manuals by LLM Agents via Interactive Environmental Learning}, 2024.

\bibitem{majumder2023clincontinuallylearninglanguage}
Bodhisattwa~Prasad Majumder, Bhavana~Dalvi Mishra, Peter Jansen, Oyvind Tafjord, Niket Tandon, Li~Zhang, Chris Callison-Burch, and Peter Clark.
\newblock {CLIN: A Continually Learning Language Agent for Rapid Task Adaptation and Generalization}, 2023.

\bibitem{nottingham2024sso}
Kolby Nottingham, Bodhisattwa~Prasad Majumder, Bhavana Dalvi~Mishra, Sameer Singh, Peter Clark, and Roy Fox.
\newblock {Skill Set Optimization: Reinforcing Language Model Behavior via Transferable Skills}.
\newblock {\em arXiv}, 2024.

\bibitem{xi2024agentgymevolvinglargelanguage}
Zhiheng Xi, Yiwen Ding, Wenxiang Chen, Boyang Hong, Honglin Guo, Junzhe Wang, Dingwen Yang, Chenyang Liao, Xin Guo, Wei He, Songyang Gao, Lu~Chen, Rui Zheng, Yicheng Zou, Tao Gui, Qi~Zhang, Xipeng Qiu, Xuanjing Huang, Zuxuan Wu, and Yu-Gang Jiang.
\newblock {AgentGym: Evolving Large Language Model-based Agents across Diverse Environments}, 2024.

\bibitem{yuan2025agentrtraininglanguagemodel}
Siyu Yuan, Zehui Chen, Zhiheng Xi, Junjie Ye, Zhengyin Du, and Jiecao Chen.
\newblock {Agent-R: Training Language Model Agents to Reflect via Iterative Self-Training}, 2025.

\bibitem{chen-etal-2024-agent}
Zehui Chen, Kuikun Liu, Qiuchen Wang, Wenwei Zhang, Jiangning Liu, Dahua Lin, Kai Chen, and Feng Zhao.
\newblock Agent-{FLAN}: Designing data and methods of effective agent tuning for large language models.
\newblock In Lun-Wei Ku, Andre Martins, and Vivek Srikumar, editors, {\em Findings of the Association for Computational Linguistics: ACL 2024}, pages 9354--9366, Bangkok, Thailand, August 2024. Association for Computational Linguistics.

\bibitem{song-etal-2024-agentbank}
Yifan Song, Weimin Xiong, Xiutian Zhao, Dawei Zhu, Wenhao Wu, Ke~Wang, Cheng Li, Wei Peng, and Sujian Li.
\newblock {A}gent{B}ank: Towards generalized {LLM} agents via fine-tuning on 50000+ interaction trajectories.
\newblock In Yaser Al-Onaizan, Mohit Bansal, and Yun-Nung Chen, editors, {\em Findings of the Association for Computational Linguistics: EMNLP 2024}, pages 2124--2141, Miami, Florida, USA, November 2024. Association for Computational Linguistics.

\bibitem{shi-etal-2024-direct}
Wentao Shi, Mengqi Yuan, Junkang Wu, Qifan Wang, and Fuli Feng.
\newblock Direct multi-turn preference optimization for language agents.
\newblock In Yaser Al-Onaizan, Mohit Bansal, and Yun-Nung Chen, editors, {\em Proceedings of the 2024 Conference on Empirical Methods in Natural Language Processing}, pages 2312--2324, Miami, Florida, USA, November 2024. Association for Computational Linguistics.

\bibitem{song2024trialerrorexplorationbasedtrajectory}
Yifan Song, Da~Yin, Xiang Yue, Jie Huang, Sujian Li, and Bill~Yuchen Lin.
\newblock {Trial and Error: Exploration-Based Trajectory Optimization for LLM Agents}, 2024.

\bibitem{yao2023tree}
Shunyu Yao, Dian Yu, Jeffrey Zhao, Izhak Shafran, Thomas~L. Griffiths, Yuan Cao, and Karthik~R Narasimhan.
\newblock {Tree of Thoughts: Deliberate Problem Solving with Large Language Models}.
\newblock In {\em Thirty-seventh Conference on Neural Information Processing Systems}, 2023.

\bibitem{aksitov2023restmeetsreactselfimprovement}
Renat Aksitov, Sobhan Miryoosefi, Zonglin Li, Daliang Li, Sheila Babayan, Kavya Kopparapu, Zachary Fisher, Ruiqi Guo, Sushant Prakash, Pranesh Srinivasan, Manzil Zaheer, Felix Yu, and Sanjiv Kumar.
\newblock {ReST meets ReAct: Self-Improvement for Multi-Step Reasoning LLM Agent}, 2023.

\bibitem{huang-etal-2023-large}
Jiaxin Huang, Shixiang Gu, Le~Hou, Yuexin Wu, Xuezhi Wang, Hongkun Yu, and Jiawei Han.
\newblock Large language models can self-improve.
\newblock In Houda Bouamor, Juan Pino, and Kalika Bali, editors, {\em Proceedings of the 2023 Conference on Empirical Methods in Natural Language Processing}, pages 1051--1068, Singapore, December 2023. Association for Computational Linguistics.

\bibitem{qu2024recursiveintrospectionteachinglanguage}
Yuxiao Qu, Tianjun Zhang, Naman Garg, and Aviral Kumar.
\newblock Recursive introspection: Teaching language model agents how to self-improve, 2024.

\bibitem{zelikman2022starbootstrapping}
Eric Zelikman, Yuhuai Wu, Jesse Mu, and Noah~D. Goodman.
\newblock {STaR: Bootstrapping Reasoning With Reasoning}, 2022.

\bibitem{zeng2023agenttuningenablinggeneralizedagent}
Aohan Zeng, Mingdao Liu, Rui Lu, Bowen Wang, Xiao Liu, Yuxiao Dong, and Jie Tang.
\newblock {AgentTuning: Enabling Generalized Agent Abilities for LLMs}, 2023.

\bibitem{prasad-etal-2024-adapt}
Archiki Prasad, Alexander Koller, Mareike Hartmann, Peter Clark, Ashish Sabharwal, Mohit Bansal, and Tushar Khot.
\newblock {AD}a{PT}: As-needed decomposition and planning with language models.
\newblock In Kevin Duh, Helena Gomez, and Steven Bethard, editors, {\em Findings of the Association for Computational Linguistics: NAACL 2024}, pages 4226--4252, Mexico City, Mexico, June 2024. Association for Computational Linguistics.

\bibitem{gpt3.5}
OpenAI.
\newblock Chatgpt blog post.
\newblock \url{https://openai.com/blog/chatgpt}, 2022.
\newblock Accessed: 2024-09-06.

\bibitem{gpt4}
OpenAI.
\newblock Gpt-4 technical report, 2023.

\bibitem{anthropic2023claude}
Anthropic.
\newblock Introducing claude 2.1, Nov 2023.
\newblock Available from Anthropic: \url{https://www.anthropic.com/news/claude-2-1}.

\bibitem{deepseekai2025deepseekv3technicalreport}
DeepSeek-AI, Aixin Liu, Bei Feng, Bing Xue, Bingxuan Wang, Bochao Wu, Chengda Lu, Chenggang Zhao, Chengqi Deng, Chenyu Zhang, Chong Ruan, Damai Dai, Daya Guo, Dejian Yang, Deli Chen, Dongjie Ji, Erhang Li, Fangyun Lin, Fucong Dai, Fuli Luo, Guangbo Hao, Guanting Chen, Guowei Li, H.~Zhang, Han Bao, Hanwei Xu, Haocheng Wang, Haowei Zhang, Honghui Ding, Huajian Xin, Huazuo Gao, Hui Li, Hui Qu, J.~L. Cai, Jian Liang, Jianzhong Guo, Jiaqi Ni, Jiashi Li, Jiawei Wang, Jin Chen, Jingchang Chen, Jingyang Yuan, Junjie Qiu, Junlong Li, Junxiao Song, Kai Dong, Kai Hu, Kaige Gao, Kang Guan, Kexin Huang, Kuai Yu, Lean Wang, Lecong Zhang, Lei Xu, Leyi Xia, Liang Zhao, Litong Wang, Liyue Zhang, Meng Li, Miaojun Wang, Mingchuan Zhang, Minghua Zhang, Minghui Tang, Mingming Li, Ning Tian, Panpan Huang, Peiyi Wang, Peng Zhang, Qiancheng Wang, Qihao Zhu, Qinyu Chen, Qiushi Du, R.~J. Chen, R.~L. Jin, Ruiqi Ge, Ruisong Zhang, Ruizhe Pan, Runji Wang, Runxin Xu, Ruoyu Zhang, Ruyi Chen, S.~S. Li, Shanghao Lu, Shangyan Zhou, Shanhuang
  Chen, Shaoqing Wu, Shengfeng Ye, Shengfeng Ye, Shirong Ma, Shiyu Wang, Shuang Zhou, Shuiping Yu, Shunfeng Zhou, Shuting Pan, T.~Wang, Tao Yun, Tian Pei, Tianyu Sun, W.~L. Xiao, Wangding Zeng, Wanjia Zhao, Wei An, Wen Liu, Wenfeng Liang, Wenjun Gao, Wenqin Yu, Wentao Zhang, X.~Q. Li, Xiangyue Jin, Xianzu Wang, Xiao Bi, Xiaodong Liu, Xiaohan Wang, Xiaojin Shen, Xiaokang Chen, Xiaokang Zhang, Xiaosha Chen, Xiaotao Nie, Xiaowen Sun, Xiaoxiang Wang, Xin Cheng, Xin Liu, Xin Xie, Xingchao Liu, Xingkai Yu, Xinnan Song, Xinxia Shan, Xinyi Zhou, Xinyu Yang, Xinyuan Li, Xuecheng Su, Xuheng Lin, Y.~K. Li, Y.~Q. Wang, Y.~X. Wei, Y.~X. Zhu, Yang Zhang, Yanhong Xu, Yanhong Xu, Yanping Huang, Yao Li, Yao Zhao, Yaofeng Sun, Yaohui Li, Yaohui Wang, Yi~Yu, Yi~Zheng, Yichao Zhang, Yifan Shi, Yiliang Xiong, Ying He, Ying Tang, Yishi Piao, Yisong Wang, Yixuan Tan, Yiyang Ma, Yiyuan Liu, Yongqiang Guo, Yu~Wu, Yuan Ou, Yuchen Zhu, Yuduan Wang, Yue Gong, Yuheng Zou, Yujia He, Yukun Zha, Yunfan Xiong, Yunxian Ma, Yuting Yan, Yuxiang
  Luo, Yuxiang You, Yuxuan Liu, Yuyang Zhou, Z.~F. Wu, Z.~Z. Ren, Zehui Ren, Zhangli Sha, Zhe Fu, Zhean Xu, Zhen Huang, Zhen Zhang, Zhenda Xie, Zhengyan Zhang, Zhewen Hao, Zhibin Gou, Zhicheng Ma, Zhigang Yan, Zhihong Shao, Zhipeng Xu, Zhiyu Wu, Zhongyu Zhang, Zhuoshu Li, Zihui Gu, Zijia Zhu, Zijun Liu, Zilin Li, Ziwei Xie, Ziyang Song, Ziyi Gao, and Zizheng Pan.
\newblock Deepseek-v3 technical report, 2025.

\bibitem{zeng-etal-2024-agenttuning}
Aohan Zeng, Mingdao Liu, Rui Lu, Bowen Wang, Xiao Liu, Yuxiao Dong, and Jie Tang.
\newblock {A}gent{T}uning: Enabling generalized agent abilities for {LLM}s.
\newblock In Lun-Wei Ku, Andre Martins, and Vivek Srikumar, editors, {\em Findings of the Association for Computational Linguistics: ACL 2024}, pages 3053--3077, Bangkok, Thailand, August 2024. Association for Computational Linguistics.

\bibitem{kingma2014adam}
Diederik~P Kingma.
\newblock {Adam: A method for stochastic optimization}.
\newblock {\em arXiv preprint arXiv:1412.6980}, 2014.

\end{thebibliography}
\bibliographystyle{unsrt}
\clearpage
\clearpage
\appendix
\label{sec:appendix}
\onecolumn

\section*{Appendix}

\section{Limitations}
\method incurs higher computational time compared to baseline methods but remains within a reasonable range. 
As shown in Table~\ref{tab:efficiency}, \method requires approximately four times the computational time of the baseline. 
However, this additional cost results in a substantial performance improvement, with \method achieving a 43.31\% gain over No Critique. 
This indicates that the extra computational time is effectively utilized. 
In contrast, other verbal-based methods, such as GPT-4o and Self-Critique, do not produce significant improvements and require more computational time than \method.

\section{Broader Impacts}
\paragraph{Positive Societal Impacts.}
By generating and leveraging natural language critiques, \method enhances the performance in embodied AI tasks (\eg, ScienceWorld) and real-world scenarios (\eg, WebShop), leading to improved task performance. 
This contributes to enabling agents to better understand and align with human goals, ultimately supporting more helpful, interpretable AI systems.  
Moreover, our critic model demonstrates strong performance despite using a relatively small backbone (Llama-3-8B), which significantly reduces the computational and environmental costs compared to larger proprietary models. 
This also promotes broader accessibility to advanced feedback mechanisms.
\paragraph{Negative Societal Impacts and Risks.}
Natural-language critiques may inadvertently incorporate biases from the training data or expose private information through the agent's reasoning process. 
Moreover, as these critiques provide explicit, step-by-step explanations, they could be misused by bad actors to facilitate disinformation, social engineering, or the generation of unsafe code. 
The relatively small size of the critic model further reduces the cost of replication and misuse, increasing the associated risks.

\section{Notations}
\begin{table}[h]
\setlength\tabcolsep{3.6pt}
\centering
\scalebox{0.9}{
\begin{tabular}{>{\raggedright\arraybackslash}p{3cm}p{12cm}}
\toprule
\multirow{2}{*}{\textbf{Symbol}} & \multicolumn{1}{c}{\textbf{Dataset Definitions}} \\
\cmidrule{2-2}
& \multicolumn{1}{c}{\textbf{Meaning}} \\
\midrule
$\mathcal{D}_{\text{critique}}$ & Expert critique dataset generated by expert critic (GPT-4o), used for training the critic model as described in \S 4.2 \\
\addlinespace
$\mathcal{D}_{\text{correct}}$ & Correct trajectory dataset collected during each iteration of the action refinement process \\
\addlinespace
$\mathcal{D}_{\text{refine}}$ & Critique-action pairs dataset used for learning how to utilize critiques effectively \\
\addlinespace
$\mathcal{D}_{\text{general}}$ & General conversational dataset (\emph{e.g.}, ShareGPT) used to maintain general language modeling capabilities during fine-tuning \\
\midrule
\midrule
\multirow{2}{*}{\textbf{Symbol}} & \multicolumn{1}{c}{\textbf{Loss Function Parameters}} \\
\cmidrule{2-2}
& \multicolumn{1}{c}{\textbf{Meaning}} \\
\midrule
$\mathcal{L}_{\text{actor}}$ & Actor model loss function with three components: learning from expert/correct trajectories, learning to utilize critiques, and maintaining general capabilities \\
\addlinespace
$\beta$ & Weight parameter balancing agentic task learning vs. general capabilities (typically $\beta = 0.8$) \\
\addlinespace
$y$ & General text responses from datasets like ShareGPT for maintaining language modeling abilities \\
\bottomrule
\end{tabular}
}
\end{table}

\clearpage
\label{app:notification}
\section{Instruction Prompt Examples}
\label{sec:critic_prompt}
The system prompts for three agentic environments are presented in Listing~\ref{listing:prompt}. 
The instruction for critique generation introduced in \S \ref{sec:critic_generation} is presented in Listing~\ref{listing:critique_generation}.
\lstset{
    backgroundcolor=\color[RGB]{245,245,244},
    breaklines=true,
    breakindent=0pt,
    basicstyle=\ttfamily\small,
    emph={ScienceWorld, WebShop, TextCraft, Instruction, Thought, Action},
    emphstyle={\bfseries\color{brown}}
}
\begin{lstlisting}[caption={Prompt details for ScienceWorld, WebShop, and TextCraft.},label=listing:prompt]

ScienceWorld Instruction:
You are an agent for the science world. Every round I will give you an observation, you have to respond with an action based on the observation to finish the given task.
Here are the actions you may take:
{"action": "open/close OBJ", "description": "open/close a container",}
{"action": "de/activate OBJ", "description": "activate/deactivate a device",}
{"action": "connect OBJ to OBJ", "description": "connect electrical components", }
{"action": "disconnect OBJ", "description": "disconnect electrical components",}
{"action": "use OBJ [on OBJ]", "description": "use a device/item",}
{"action": "look around", "description": "describe the current room",}
{"action": "look at OBJ", "description": "describe an object in detail",}
{"action": "look in OBJ", "description": "describe a container's contents",}
{"action": "read OBJ", "description": "read a note or book",}
{"action": "move OBJ to OBJ", "description": "move an object to a container", }
{"action": "pick up OBJ", "description": "move an object to the inventory", }
{"action": "put down OBJ", "description": "drop an inventory item",}
{"action": "pour OBJ into OBJ", "description": "pour a liquid into a container", }
{"action": "dunk OBJ into OBJ", "description": "dunk a container into a liquid", }
{"action": "mix OBJ", "description": "chemically mix a container",}
{"action": "go to LOC", "description": "move to a new location",}
{"action": "eat OBJ", "description": "eat a food",}
{"action": "flush OBJ", "description": "flush a toilet",}
{"action": "focus on OBJ", "description": "signal intent on a task object",}
{"action": "wait", "description": "take no action for 10 iterations",}
{"action": "wait1", "description": "take no action for 1 iteration", }
{"action": "task", "description": "describe current task",}
{"action": "inventory", "description": "list your inventory"}
Your response should use the following format:

Thought: your thoughts.
Action: your next action

WebShop Instruction:
You are web shopping. I will give you instructions about what to do. You have to follow the instructions. Every round I will give you an observation and a list of available actions, you have to respond an action based on the state and instruction.
You can use search action if search is available. You can click one of the buttons in clickables. An action should be of the following structure: search[keywords] click[value]. If the action is not valid, perform nothing. Keywords in search are up to you, but the value in click must be a value in the list of available actions. Remember that your keywords in search should be carefully designed. Your response should use the following format:

Thought: I think ... 
Action: click[something] or search[something]

TextCraft Instruction:
You are given a few useful crafting recipes to craft items in Minecraft. Crafting commands are of the format "craft [target object] using [input ingredients]". Every round I will give you an observation, you have to respond to an action based on the state and instruction. You can "get" an object (ingredients) from the inventory or the environment, look up the game "inventory" by inventory, or "craft" (target) using any of the crafting commands. You can use ONLY these crafting commands provided, do not use your own crafting commands. However, if the crafting command uses a generic ingredient like "planks", you can use special types of the same ingredient e.g. dark oak "planks" in the command instead. Your response should use the following format:

Thought: ...
Action: ...
\end{lstlisting}
\lstset{
    backgroundcolor=\color[RGB]{245,245,244},
    breaklines=true,
    breakindent=0pt,
    basicstyle=\ttfamily\small,
    emph={Critique, Generation, Instruction, Steps, Dimensions, Format, Inputs, Step, 1, 2},
    emphstyle={\bfseries\color{brown}}
}
\begin{lstlisting}[caption={The instruction prompts for Critique Generation.},label=listing:critique_generation]
Critique Generation Instruction:
Your task is to critique the candidate's next-step action based on the agent's task goal and interaction history.

{available_actions}

Critique Steps
Step 1: Analyze Candidate Action
Examine the candidate's action based on the following criteria and assign an overall grade using this scale: Excellent, Good, Neutral, Poor, Very Poor.

Critique Dimensions
- Contribution: Assess whether the action contributes to completing the agent's task. This includes both direct actions (e.g., picking up the target OBJ) and indirect actions (e.g., reasonable exploration that can provide additional environmental information and facilitate future progress).
- Feasibility: Assess whether the action is valid according to the agent's predefined Allowed Action Types list.
- Efficiency: Analyze whether the action optimally achieves the task without unnecessary steps or redundancy.

Step 2: Provide Revision Suggestions
Suggest a modification to align the candidate's action better with the task or the agent's action capabilities. Note that the suggested revision should be based on the Allowed Action and Object Types.

Critique Format
Please structure your critique in the following format:
## Contribution: [Analysis of Contribution].
## Feasibility: [Analysis of feasibility].
## Efficiency: [Analysis of efficiency].
## Overall Grading: [Overall grade: Excellent/Good/Neutral/Poor/Very Poor].
## Suggested Revision: [Brief revision suggestion, if applicable].

Inputs:
The agent's task goal and interaction history:

{history}

Candidate next step action: {candidate_action}

Now, please provide your critique:

Critique Generation Instruction (with expert path):
Your task is to critique the candidate next-step action based on the agent's task goal and interaction history. The gold path for current task is provided as a reference to guide your critique, but do not explicitly mention it in your critique.

{available_actions}

Critique Steps

Step 1: Analyze Candidate Action
Examine the candidate's action based on the following criteria, then assign an overall grade using this scale: Excellent, Good, Neutral, Poor, Very Poor.

Critique Dimensions
-Contribution: Assess whether the action contributes to completing the agent's task. This includes both direct actions (e.g., picking up the target OBJ) and indirect actions (e.g., reasonable exploration that can provide additional environmental information and facilitate future progress).
-Feasibility: Assess whether the action is valid according to the agent's predefined Allowed Action Types list.
-Efficiency: Analyze whether the action optimally achieves the task without unnecessary steps or redundancy.

Step 2: Provide Revision Suggestions
Suggest a modification to align the candidate's action better with the task or the agent's action capabilities. For example, if the action is not allowed, recommend an alternative from the action list that aligns better with the task goal.

Critique Format
Please structure your critique in the following format:
## Contribution: [Analysis of Contribution].
## Feasibility: [Analysis of feasibility].
## Efficiency: [Analysis of efficiency].
## Overall Grading: [Overall grade: Excellent/Good/Neutral/Poor/Very Poor].
## Suggested Revision: [Brief revision suggestion, if applicable].

Referenced Gold Path for Current Task:

{gold_path}

Inputs:
The agent's task goal and interaction history:

{history}

Candidate next step action: {candidate_action}

Now, please provide your critique:

\end{lstlisting}

\clearpage

\section{Evaluation Settings}
In the experiments presented in this paper, we used three agentic environments: WebShop, SciWorld, and TextCraft.
\label{sec:eval_settings}

\subsection{Details of Environments}
\paragraph{WebShop}
WebShop is an interactive web environment designed for web shopping. 
In this environment, agents are given instructions and must purchase a product that meets specified criteria. 
Agents can either click a button on the webpage or use the search engine to find the product. 
We evaluate performance using the success rate, with a maximum of 10 rounds per task.

\paragraph{ScienceWorld} 
ScienceWorld is a benchmark environment for testing agents’ scientific reasoning abilities, based on a standard elementary science curriculum. 
It includes 30 types of tasks, such as using measurement instruments and conducting mechanics experiments. 
The action space is task-specific, with the environment simulator providing the effects of actions. 
We use reward as the evaluation metric, with a maximum of 30 rounds per task.
Task lengths are classified based on the ScienceWorld environment, which provides gold-standard trajectories from 30 hand-coded oracle agents. 
We adopt the following categorization: Short (11.76), Medium (28.58), and Long (94.30). 
These lengths correspond to the average number of steps taken by the oracle agent.

\paragraph{TextCraft} TextCraft is a text-only environment for crafting Minecraft items. 
It constructs a crafting tree based on Minecraft’s crafting recipes, consisting of 544 nodes, each representing a target item. For each task, the agent is given a target item and a list of crafting commands generated by the tree. 
Tasks vary in complexity, ranging from 1 to 4 steps. 
The environment supports three valid actions:``craft <item> using <ingredients>'', ``get <item>'', and ``inventory''. After each round, the environment checks the agent’s actions and returns the execution state. 
Agents are rewarded with 1 point only upon successfully crafting the target item. We evaluate performance using the success rate, with a maximum of 20 rounds per task.

\subsection{Computational Efficiency Analysis}
\label{app:resource}
We evaluated the computational efficiency of \method by conducting experiments in three agentic environments, using Llama-3-8B as the actor model. 
We measured the average inference time per task. 
As shown in Table~\ref{tab:efficiency}, although \method incurred higher computation times compared to inference without critique guidance (i.e., No Critique), these times remained within an acceptable range.
Furthermore, \method demonstrated a substantial reduction in inference time compared to using GPT-4 as the critic, while achieving significant performance improvements.
\begin{table*}[t!]
\setlength\tabcolsep{3pt}
\caption{Computational efficiency of different methods in three agentic environments per task.}
\scalebox{0.97}[0.97]{ 
\small
  \centering
    \begin{tabular}{lcccccc}
    \toprule
    \multirow{2}{*}{\textbf{Method}} & \multicolumn{2}{c}{\textbf{WebShop}} & \multicolumn{2}{c}{\textbf{Sciworld}} & \multicolumn{2}{c}{\textbf{Textcraft}} \\
    \cmidrule(lr){2-3} \cmidrule(lr){4-5} \cmidrule(lr){6-7}
    & \multicolumn{1}{c}{Inference Time} 
    & \multicolumn{1}{c}{Performance} 
    & \multicolumn{1}{c}{Inference Time} 
    & \multicolumn{1}{c}{Performance} 
    & \multicolumn{1}{c}{Inference Time}  
    & \multicolumn{1}{c}{Performance}  \\
    \midrule
    No Critique    & 0.38 min & 13.49 & 1.49 min & 14.48  & 1.02 min & 10.00 \\
    \ w/ DGAP & 1.13 min & 30.41  & 3.13 min & 19.52  & 1.33 min & 21.00   \\
    \ w/ Self-Critique  &1.36 min  & 1.50  & 3.42min & 10.06  & 2.17 min  & 19.00  \\
    \ w/ GPT-4o  &2.34 min  & 17.78  & 6.28 min & 33.06  & 3.28 min & 46.00   \\
    \ w/ Critic Model  &1.22 min  & \textbf{56.80} & 3.41 min  & \textbf{68.51}  & 1.98 min & \textbf{59.00}   \\
    \bottomrule
    \end{tabular}
}
  \label{tab:efficiency}
\end{table*}

\clearpage
\section{Implementation Details}
\label{sec:implementation_details}
In our experiments, we use the Llama-3-8B-Instruct~\citep{llama3modelcard} as the base model for both the actor and the critic.

\subsection{Training Data} \paragraph{Critique Generation} 
The training data for the critic model consists of expert critiques generated by the expert critic (\ie GPT-4o) in the ScienceWorld, Webshop, and TextCraft environments, as described in \S~\ref{sec:critic_generation}. 
The specific training set sizes during the SFT phases for Llama3-8B-Instruct are 14K from ScienceWorld, 10K from Webshop, and 8K from TextCraft. Examples of our SFT training data are provided below:
\lstset{
    backgroundcolor=\color[RGB]{245,245,244},
    breaklines=true,
    breakindent=0pt,
    basicstyle=\ttfamily\small,
    emph={Example, conversations, human, gpt, from, value, instruction, system, SFT},
    emphstyle={\bfseries\color{brown}}
}
\begin{lstlisting}[label=sft:example]
SFT Example: 
{"conversations": 
[{"from": "human", "value": "Your task is to critique the candidate's next-step action based on the agent's task goal and interaction history. Allowed Action Types:- search[KEYWORDS]: perform a web search - click[OBJ]: click an object or link on a webpage. Step 1: Analyze Candidate Action.Examine the candidate's action based on the following criteria and assign an overall grade using this scale: Excellent, Good, Neutral, Poor, Very Poor. Critique Dimensions: Contribution: Assess whether the action contributes to completing the agent's task. This includes both direct actions and indirect actions. Feasibility: Assess whether the action is valid according to the agent's predefined action list. Specifically, check if the action is part of the Allowed Action Types, whether the OBJ/KEYWORDS within the action is clearly specified, and whether it is formatted correctly according to the system's requirements. Efficiency: Analyze whether the action optimally achieves the task without unnecessary steps or redundancy. Step 2: Provide Revision Suggestionn: Suggest a modification to align the candidate's action better with the task or the agent's action capabilities. For example, if the action is not allowed, recommend an alternative from the action list that aligns better with the task goal. Critique Format:Please structure your critique in the following format : ## Contribution: [Analysis of Contribution]. ## Feasibility: [Analysis of feasibility]. ## Efficiency: [Analysis of efficiency].## Overall Grading: [Overall grade: Excellent/Good/Neutral/Poor/Very Poor]. ## Suggested Revision: [Brief revision suggestion, if applicable]. ### InputsThe Agent's Task Goal and Interaction History: WebShop [SEP] Instruction: [SEP] Find me machine wash, moisture wicking men's t-shirts & tanks with polyester spandex [SEP] Search Candidate Next Step Action: search[machine wash moisture wicking men's t-shirts with polyester spandex long sleeve red blue XXL price: under 50]. Now, please provide your critique:"},
{"from": "gpt", "value": "## Contribution: The candidate's action contributes to the task by attempting to perform a search for the specified product. It includes most of the necessary attributes such as material, sleeve length, color, size, and price, which are crucial for finding the desired item. ## Feasibility: The action is feasible as it uses the allowed action type "search[KEYWORDS]". However, there are some discrepancies in the keywords used. The candidate action uses "XXL" instead of "xx-large" and "price: under 50" instead of "price < 50.00". While these variations might still yield relevant results, they do not exactly match the specified format. ## Efficiency: The action is relatively efficient as it attempts to gather relevant search results in one step. However, the slight variations in keyword formatting could potentially lead to less precise search results, which might require additional steps to refine the search. ## Overall Grading: Good. ## Suggested Revision: Modify the search keywords to exactly match the specified format: search[machine wash moisture wicking men's t-shirts & tanks polyester spandex long sleeve red blue size xx-large price < 50.00]. This ensures consistency and potentially more accurate search results."}], 
"system": ""}}

\end{lstlisting}

\paragraph{Action Refinement}
As described in \S~\ref{sec:critic_utilization}, we performed action refinement three times. Each training dataset consists of two parts: Correct Path, which includes the correct trajectories obtained by the actor from interactions with the environment in the previous iteration; and Correct Critique, which includes the <critique, action> pairs that the actor correctly followed during the previous iteration. The dataset size for each iteration of Llama-3-8B-Instruct is detailed in Table~\ref{tab:datasize}.
\begin{table*}[t!]
\setlength\tabcolsep{9pt}
\small
  \centering
  \caption{Training dataset size for each iteration of Action Refinement stage.}
    \begin{tabular}{lcccccccc}
    \toprule
    \multirow{2}{*}{\textbf{Env}} & \multicolumn{2}{c}{\textbf{Gold}} & \multicolumn{2}{c}{\textbf{\#Iter 1}} & \multicolumn{2}{c}{\textbf{\#Iter 2}}
    & \multicolumn{2}{c}{\textbf{\#Iter 3}} \\
    \cmidrule(lr){2-3} \cmidrule(lr){4-5} \cmidrule(lr){6-7} \cmidrule(lr){8-9}
    & \multicolumn{1}{c}{Path} 
    & \multicolumn{1}{c}{Critique} 
    & \multicolumn{1}{c}{Path} 
    & \multicolumn{1}{c}{Critique}
    & \multicolumn{1}{c}{Path} 
    & \multicolumn{1}{c}{Critique} 
    & \multicolumn{1}{c}{Path} 
    & \multicolumn{1}{c}{Critique} \\
    \midrule
    \textbf{ScienceWorld}   & 378 & 4106    & 133 & 1464    & 158 & 1803 & 193 & 2205    \\
    \textbf{WebShop} & 427 & 1659  & 106 & 663  & 283 & 1637 & 344 & 2010  \\
    \textbf{TextCract}  &441  & 1701  & 233 & 1423  & 310 & 2070 & 340 & 2049  \\
    \textbf{OverAll}  &1246  & 6740  & 472 & 3550  & 751 & 5510 & 877 & 6264  \\
    \bottomrule
    \end{tabular}
  \label{tab:datasize}
\end{table*}

\subsection{Finetuning Details}
We ran SFT experiments using 8 NVIDIA A100-40GB GPUs. For action refinement, to prevent policy drift, we only trained the base model (\ie, Llama-3-8B-Instruct) each time. 
We conduct experiments with the LlamaFactory code base\footnote{\url{https://github.com/hiyouga/LLaMA-Factory}}. The configurations of our hyper-parameters are detailed in Table~\ref{tab:configuration}.

\begin{table}[ht]
    \centering
    \setlength\tabcolsep{4pt}
    \caption{Fine-tuning hyper-parameters for Critique Generation and Action Reinement stage.}
    \begin{tabular}{lcc}
    \toprule
    Configuration & Critique Generation & Action Refinement \\
    \midrule
    Model & Llama-3-8B-Instruct  & Llama-3-8B-Instruct \\
    Number of epochs & 3 & 3 \\
    Devices &8 A100 GPU $(40 \mathrm{~GB})$ &8 A100 GPU $(40 \mathrm{~GB})$\\
    Total Batch size & 64 samples & 64 samples \\
    Optimizer & Adam~\cite{kingma2014adam} & Adam~\cite{kingma2014adam} \\
    & $(\beta_1=0.9, \beta_2=0.98, \epsilon=1 \times 10^{-8})$ & $(\beta_1=0.9, \beta_2=0.98, \epsilon=1 \times 10^{-8})$\\
    Learning rate & $2 \times 10^{-5}$ & $2 \times 10^{-5}$ \\
    Warmup Ratio & 0.05 & 0.05 \\
    Cutoff Length & 4096 & 4096 \\
    Training Time &5h 46m 24s &11h 11m 55s \\
    \bottomrule
    \end{tabular}
    \vspace{0.15cm}
    \label{tab:configuration}
\end{table}

\subsection{Effect of Iteration Count}
\begin{wrapfigure}{r}{0.46\textwidth}
    \centering
    \vspace{-1.3cm}
    \includegraphics[width=\linewidth]{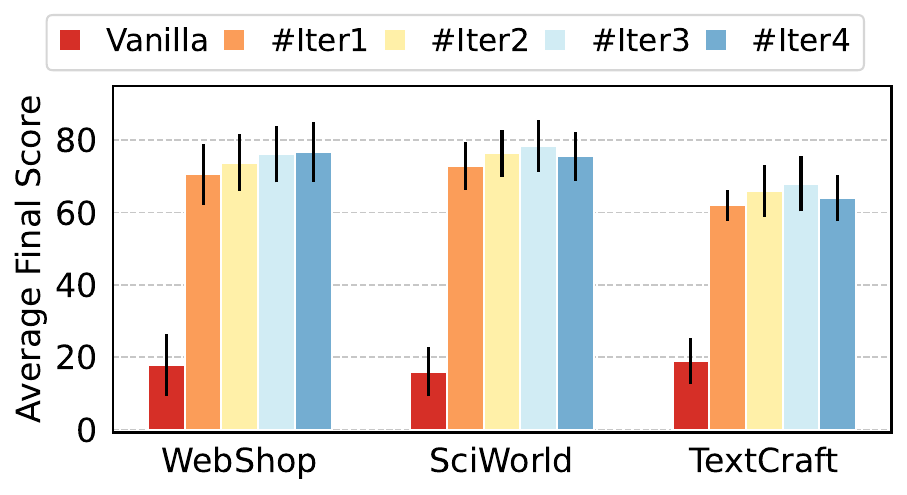}
    \vspace{-0.61cm} 
    \caption{Performance of models with varying iteration counts across three tasks.}
    \label{fig:app_01}
\end{wrapfigure}
Here we analyze the impact of the number of iterations on the performance of \method. 
As shown in Figure~\ref{fig:app_01}, the fourth iteration leads to a performance drop in ScienceWorld and TextCraft, while the improvement in WebShop is marginal (+0.67\%). The third iteration achieves the highest average performance across all tasks. These results suggest that additional iterations do not yield further gains. Therefore, we report the results from the third iteration in the main results section (\S~\ref{sec:main_results}).

\subsection{Effect of Training with Expert Trajectories and Updating the Critic}
\label{app:updating_results}
\begin{table*}[h]
\setlength\tabcolsep{11pt}
\small
\centering
\caption{Performance comparison across different methods and datasets}
\begin{tabular}{lcccc}
\toprule
\textbf{Method} & \textbf{WebShop} & \textbf{ScienceWorld} & \textbf{TextCraft} & \textbf{Average} \\
\midrule
Llama3-8B & 13.49 & 14.48 & 10.00 & 12.66 \\
w/ GPT-4o & 17.78 & 33.06 & 46.00 & 32.28 \\
w/ Critic (Trained on GPT-4o Traj) & 43.21 & 56.93 & 52.00 & 50.71 \\
w/ Critic (Trained on Expert Traj) & 56.80 & 68.51 & 59.00 & 61.44 \\
w/ Updated Critic & 62.51 & 74.56 & 64.00 & 67.02 \\
\bottomrule
\end{tabular}
\label{tab:updating_results}
\end{table*}
In this section, we analyze two key factors: 
\begin{inparaenum}[\it 1)] 
\item whether our critic heavily depends on expert trajectories;
\item whether updating the critic model through iterations can continue to improve performance. 
\end{inparaenum}
To evaluate the first factor, we replace the gold-standard expert trajectories with GPT-4o-generated trajectories across three environments, following the exact pipeline described in \S \ref{sec:critic_generation} for training the critic model. 
For the second factor, we perform iterative updates on the critic model after action refinement to adapt to the updated actor's policy. Specifically, we use the critic from the first iteration to generate new critique data and retrain the critic.

As shown in Table~\ref{tab:updating_results}, CGI remains highly effective even without gold-standard expert data. Our method achieves an average performance of 50.71, significantly outperforming both the baseline and the GPT-4o model that generated the initial data. This highlights the robustness of our framework. 
Additionally, the critic can be dynamically updated. The results indicate that updating the critic leads to further performance improvements (from 61.44\% to 67.02\%), demonstrating that the critic is not limited to being static and can evolve alongside the actor.

\section{Baselines}
\label{sec:app_baselines}
\subsection{Critique Methods}
In Section~\ref{sec:main_results}, we compare our trained Critic-Llama3 model with other verbal- and numerical-based methods. For the verbal-based approaches, such as self-critique and GPT-4, we generate critiques by prompting the model with the instructions provided in Appendix~\ref{sec:critic_prompt}. For the numerical-based method (i.e., DGAP), following ~\citep{qian2025discriminatorguided}, we compute the cosine similarity between the actor model (Llama-3-8B) and expert data at each step. We collected 23K data points for Scienceworld, 16K for WebShop, and 12K for TextCraft. Examples of the reward model training data are provided below:
\lstset{
    backgroundcolor=\color[RGB]{245,245,244},
    breaklines=true,
    breakindent=0pt,
    basicstyle=\ttfamily\small,
    emph={Example, conversations, human, gpt, from, value, instruction, input, chosen, rejected, system, Action, Score, DGAP},
    emphstyle={\bfseries\color{brown}}
}
\begin{lstlisting}[label=rm:example]
DGAP Example: 
{"input": "Your task is to measure the melting point of lead, which is located around the kitchen. 10. look around. Action: go to hallway", "Score": "10"}
{"input": "Your task is to measure the melting point of lead, which is located around the kitchen. 10. look around. Action: look at art studio", "Score": "0"}
{"input": "Your task is to measure the melting point of lead, which is located around the kitchen. 10. look around. 9. go to hallway. Action: put down orange", "Score": "0"}
{"input": "Your task is to measure the melting point of lead, which is located around the kitchen. 10. look around. Action: look at hallway", "Score": "9.03"}
{"input": "Your task is to measure the melting point of lead, which is located around the kitchen. 10. look around. 9. look at hallway. Action: open door to outside", "Score": "6.13"}
{"input": "Your task is to measure the melting point of lead, which is located around the kitchen. 10. look around. 9. look at hallway. 8. open door to outside. Action: teleport to kitchen", "Score": "8.87"}
\end{lstlisting}

\subsection{Iterative Methods}
In Section~\ref{sec:main_results}, we compare \method with two iterative baselines: Reflexion and Iterative SFT. Following \citep{shinn2023reflexion}, we prompt the model for self-reflection (Listing~\ref{listing:reflexion_instruction}) at the end of each iteration. 
The insights gained from the current round are then incorporated into the system prompt to guide the model’s decision-making in the subsequent round.
For Iterative SFT, we use the actor model $\pi_{\theta}^{k}$ from the previous round to interact with the environment. 
The newly collected correct trajectories are combined with the expert path to form the training data for the next round. Specifically, for iterations 1 to 3, we used 1,676, 1,719, and 1,750 trajectories, respectively, mixing them with the general dataset (\ie ShareGPT) at a 1:4 ratio for training.
To prevent policy drift, we update $\pi_{\theta}^{0}$ at each iteration. Each iteration consists of 3 epochs. 
The learning rate for Iterative SFT is set to 2e-5, with a 3\% warm-up and a cosine scheduler. 
We use the AdamW optimizer with a weight decay of 0 and apply a maximum gradient clipping norm of 1 to prevent gradient explosion. 
The batch size is set to 2 per device, with a sequence length of 4,096 and an accumulation count of 4, adjusted based on sequence parallelism.
\lstset{
    backgroundcolor=\color[RGB]{245,245,244},
    breaklines=true,
    breakindent=0pt,
    basicstyle=\ttfamily\small,
    emph={Reflexion, Instruction},
    emphstyle={\bfseries\color{brown}},
    caption={The instruction prompt for Reflexion.}
}
\begin{lstlisting}[label=listing:reflexion_instruction]
Reflexion Instruction:
Your task is to provide a next plan of action based on the previous attempt's interaction history. The next plan should reflect the lessons learned from the failure and adjust for better results.

Previous Trial:
{history}

Next Plan:
\end{lstlisting}

\section{Case Study}
\subsection{Failure Cases of Fine-tuned Llama3}
\label{app:sft_sample}
We provide negative examples where the fine-tuned Llama3-8B does not follow the critique.
\lstset{
    backgroundcolor=\color[RGB]{245,245,244},
    breaklines=true,
    breakindent=0pt,
    basicstyle=\ttfamily\small,
    emph={Good, Case, Of, ScienceWorld, Task, Description, Action, Observation, Critique},
    emphstyle={\bfseries\color{brown}},
    caption={Failure Cases of Fine-Tuned Llama3-8B in ScienceWorld}
}
\begin{lstlisting}[label=listing:sci_example]
Task Description: 
Your task is to measure the temperature of unknown substance B, which is located around the bathroom. First, focus on the thermometer. Next, focus on the unknown substance B. If the unknown substance B temperature is above 100.0 degrees celsius, place it in the orange box. If the unknown substance B temperature is below 100.0 degrees celsius, place it in the yellow box. The boxes are located around the bathroom.
History:
Action: open door to kitchen
Observation: The door is now open.
Action: go to kitchen
Observation: You move to the kitchen.
Action: open door to hallway
Observation: The door is now open
Action: go to hallway
Observation: You move to the hallway.
Critique:
......
### Feasibility: The candidate action "open door to bathroom" is efficient in terms of task progression, as it directly addresses the need to access the bathroom. However, the agent should consider checking the current state first to ensure the action is feasible by "Look Around".
### Overall Grading: Netual
......
Action: go to bathroom
Observation: No known action matches that input.
\end{lstlisting}

\subsection{Critic-Llama3 Outperforms GPT-4o}
\label{app:critic_examples}
We provide the following examples to explain why our critic outperforms GPT-4o when the actor model is fixed as Llama3-8B. Compared to GPT-4o, our critic demonstrates superior spatial reasoning and a higher degree of goal-directed behavior. 
\lstset{
    backgroundcolor=\color[RGB]{245,245,244},
    breaklines=true,
    breakindent=0pt,
    basicstyle=\ttfamily\small,
    caption={Comparison between Critic-Llama3 and GPT-4o as Critics},
    emph={},
    emphstyle={\bfseries},
    moredelim=[is][\color{brown}\bfseries]{@}{@}, 
}
\begin{lstlisting}[label=listing:critic_example]
@Example 1@: Spatial Reasoning
State: Agent is in the @workshop@. Task requires an object in the @bathroom@.
Action: @focus on thermometer@
GPT-4o (Incorrect): @Overall Grading: Excellent@. 
It approves the action based on the task description, ignoring the agent's location.
Our Critic (Correct): @Overall Grading: Very Poor@. 
It correctly identifies the thermometer is not in the current room and suggests navigating first.

@Example 2@: Goal-Directedness
Task: Find a living thing.
Action: @move to the closet@ (an exploratory but inefficient action).
GPT-4o (Inefficient): @Overall Grading: Neutral@. 
It allows this exploratory step.
Our Critic (Efficient): @Overall Grading: Poor@. 
It rejects the inefficient action and suggests a more direct one, like opening the door to explore other rooms.
\end{lstlisting}

\subsection{Trajectory Visualization of ScienceWorld}
\begin{figure*}[t]
    \centering
    \includegraphics[width=0.9\linewidth]{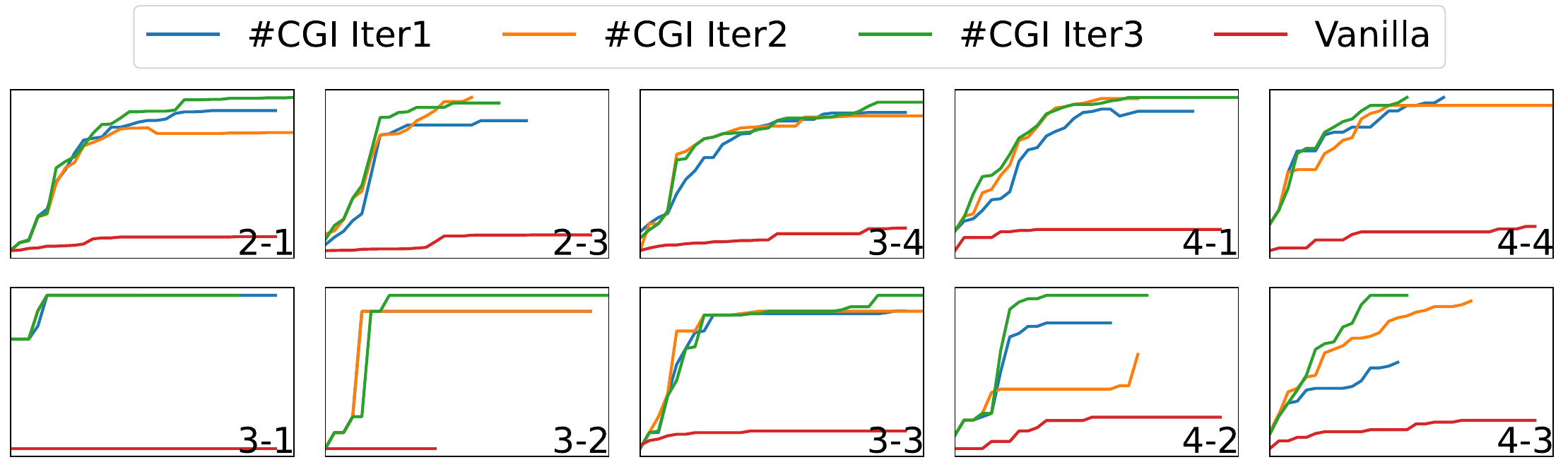}
    \caption{Visualizing trajectories of \method, and base model (Llama-3-8B-Instructt), the x-axis represents time steps, ranging from 0 to T, while the y-axis denotes scores, which vary from 0 to 100. Each graph illustrates the trajectories corresponding to different tasks in test variation. Task identifiers are positioned at the bottom right of each figure.}
    \label{fig:app_02}
\end{figure*}

We visualize the cumulative scores of the actor performing different numbers of action refinements (\#Iter $k$) and the base model in the ScienceWorld scene\footnote{Detailed information of each task can be found in \url{https://github.com/allenai/ScienceWorld}}. As shown in Figure~\ref{fig:app_02}, our \method method shows a notable improvement in efficiency, achieving higher scores in fewer steps. Furthermore, as the number of action refinements increases, efficiency also improves.

\end{document}